\documentclass[11pt]{article}

\usepackage[preprint]{acl}

\usepackage{times}
\usepackage{latexsym}
\usepackage{url}
\usepackage[T1]{fontenc}

\usepackage[utf8]{inputenc}
\usepackage{amsmath}
\usepackage{microtype}

\usepackage{inconsolata}
\usepackage{booktabs}
\usepackage{graphicx}
\usepackage{multirow}
\usepackage{xcolor}

\usepackage{tikz}
\usepackage{graphicx}
\usepackage{microtype}

\usepackage{listings}
\lstdefinestyle{prompt}{basicstyle=\ttfamily\scriptsize, breaklines=true, breakatwhitespace=true,
  columns=fullflexible, keepspaces=true, frame=single, framesep=4pt, xleftmargin=4pt, xrightmargin=4pt,
  showstringspaces=false, upquote=true}

\definecolor{kh}{HTML}{168aff}

\definecolor{nk}{HTML}{B22222}

\definecolor{justin}{HTML}{B22222}

\providecommand{\fm}{non-assertive}
\providecommand{\Fm}{Non-assertive}
\providecommand{\bench}{NAME-Bench}

\newcommand\nnfootnote[1]{%
  \begin{NoHyper}
  \renewcommand\thefootnote{}\footnote{#1}%
  \addtocounter{footnote}{-1}%
  \end{NoHyper}
}

\title{Not Every Mention Is a Claim: Rethinking Object Hallucination Detection in Vision-Language Models}

\author{Hsin-Ling Hsu$^{*,1}\quad$Neeraja Kirtane$^{*,2}\quad$Kuan-Hao Huang$^2$ \vspace{0.1em}\\
  $^1$National Chengchi University$\quad^2$Texas A\&M University\vspace{0.1em}\\
  \texttt{112306092@nccu.edu.tw} \ \ \ 
  \texttt{\{neeraja,khhuang\}@tamu.edu}}

\begin{document}
\maketitle

\begin{abstract}
Object hallucination in vision-language models (VLMs) is usually measured by checking each object a caption names against the image. Most benchmarks and detectors alike treat every object mention as a claim that the object is present. This assumption fails when a caption names an object without asserting it, as in ``a bench for \emph{people} to sit on'', which mentions people without claiming that any are in the image. We call such mentions \emph{non-assertive}. Treating them as hallucinations inflates the error rates attributed to models that use such language. To separate them from real hallucinations, we introduce \textbf{NAME-Bench} (Non-Assertive Mention Evaluation Benchmark), 25{,}497 object mentions from captions that five VLMs generate for COCO images. Each mention is labelled concrete (asserted and present), hallucinated (asserted and absent), or non-assertive, and non-assertive mentions are further divided into four subtypes: depictive, purpose, negation and hedged. All the hallucination detectors that we evaluate flag non-assertive mentions as hallucinations, at rates between 14\% and 80\%. Adding non-assertive mentions to detector training as negatives largely corrects this on LLaVA-1.5. On the four newer VLMs, a comparable correction takes five to ten times more data, and the detectors then catch fewer real hallucinations. These findings show that current detectors conflate mentioning an object with asserting it, and that this conflation is harder to undo on the four newer VLMs than on LLaVA-1.5. Hallucination evaluation should therefore separate what a caption asserts from what it merely mentions.
\end{abstract}

\section{Introduction}
\nnfootnote{$^{*}$ denotes equal contributions.}
\newcommand{\figonepath}{figures/}
\providecommand{\figonepath}{}

\definecolor{fOneGreenDark}{RGB}{16,118,60}
\definecolor{fOneGreenFill}{RGB}{46,169,106}
\definecolor{fOneGreenLine}{RGB}{120,185,155}
\definecolor{fOneGreenBg}{RGB}{242,252,247}
\definecolor{fOneRedDark}{RGB}{180,16,18}
\definecolor{fOneRedFill}{RGB}{234,72,34}
\definecolor{fOneRedLine}{RGB}{240,150,155}
\definecolor{fOneRedBg}{RGB}{254,242,242}
\definecolor{fOneOrangeDark}{RGB}{217,153,25}
\definecolor{fOneOrangeFill}{RGB}{253,174,21}
\definecolor{fOneOrangeLine}{RGB}{245,205,85}
\definecolor{fOneOrangeBg}{RGB}{253,249,236}
\definecolor{fOneTrack}{RGB}{238,238,238}

\newcommand{\fOneThr}{0.60}

\newcommand{\fOnePanel}[9]{%
  \begin{scope}[shift={#1}]
    \draw[draw=fOne#2Line, fill=fOne#2Bg, line width=0.6pt, rounded corners=4pt]
      (0,0) rectangle (636,391);
    \node[font=\bfseries\fontsize{10}{12}\selectfont, text=fOne#2Dark] at (318,31) {#3};
    \node[font=\fontsize{7.5}{9}\selectfont] at (318,68) {#4};
    \node[anchor=north west, inner sep=0pt] at (20,91)
      {\includegraphics[width=74.4pt,height=44.6pt]{\figonepath #5}};
    \node[anchor=west, align=left, inner sep=0pt,
          font=\fontsize{6.9}{8.2}\selectfont] at (352,184) {#6};
    \fill[fOneTrack] (72,297) rectangle (575,319);
    \fill[fOne#2Fill] (72,297) rectangle ({72+503*#7},319);
    \draw[densely dashed, line width=0.45pt] ({72+503*\fOneThr},288) -- ({72+503*\fOneThr},342);
    \node[anchor=base west, inner sep=0pt, font=\itshape\fontsize{5.6}{6}\selectfont]
      at ({72+503*\fOneThr-2},283) {};
    \draw[fill=white, line width=0.9pt] ({72+503*#8},308) circle (2.9pt);
    \node[font=\fontsize{7.2}{8}\selectfont] at (72,347) {0};
    \node[font=\fontsize{7.2}{8}\selectfont] at (575,347) {1};
    \node[font=\bfseries\fontsize{8.4}{10}\selectfont, text=fOne#2Dark] at (318,359) {#9};
  \end{scope}%
}

\newcommand{\fOneKw}[2]{\textbf{\textcolor{#1}{#2}}}

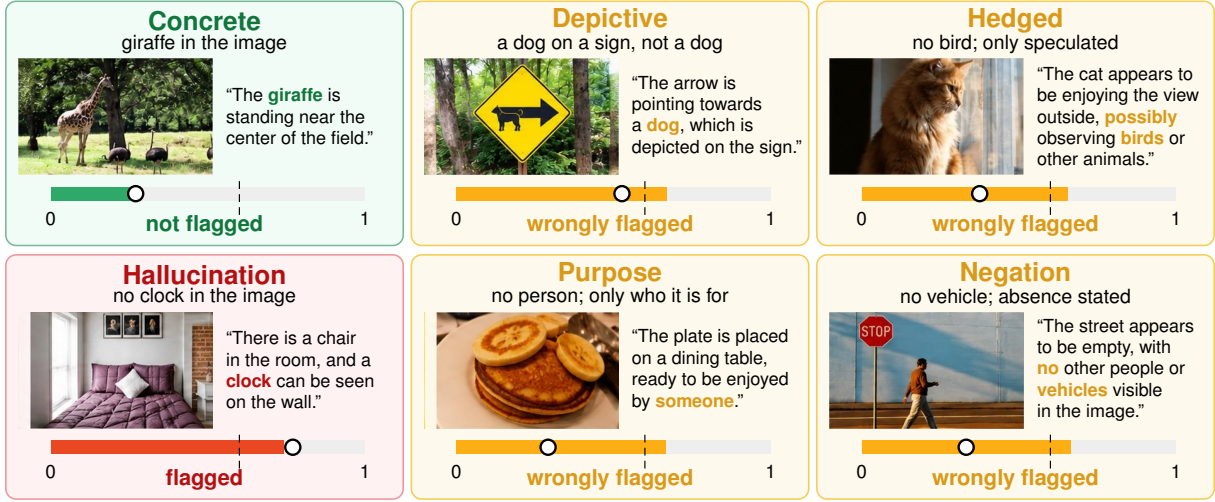
\begin{figure*}[t]
\centering
\resizebox{\textwidth}{!}{%
\fontfamily{phv}\selectfont
\begin{tikzpicture}[x=0.24pt, y=-0.24pt]
  \fOnePanel{(3,4)}{Green}{Concrete}{giraffe in the image}{img_concrete}
    {``The \fOneKw{fOneGreenDark}{giraffe} is\\standing near the\\center of the field.''}
    {0.250}{0.270}{not flagged}
  \fOnePanel{(652,4)}{Orange}{Depictive}{a dog on a sign, not a dog}{img_depictive}
    {``The arrow is\\pointing towards\\a \fOneKw{fOneOrangeDark}{dog}, which is\\depicted on the sign.''}
    {0.670}{0.527}{wrongly flagged}
  \fOnePanel{(1301,4)}{Orange}{Hedged}{no bird; only speculated}{img_hedged}
    {``The cat appears to\\be enjoying the view\\outside, \fOneKw{fOneOrangeDark}{possibly}\\observing \fOneKw{fOneOrangeDark}{birds} or\\other animals.''}
    {0.655}{0.375}{wrongly flagged}
  \fOnePanel{(3,410)}{Red}{Hallucination}{no clock in the image}{img_hallucination}
    {``There is a chair\\in the room, and a\\\fOneKw{fOneRedDark}{clock} can be seen\\on the wall.''}
    {0.740}{0.770}{flagged}
  \fOnePanel{(652,410)}{Orange}{Purpose}{no person; only who it is for}{img_purpose}
    {``The plate is placed\\on a dining table,\\ready to be enjoyed\\by \fOneKw{fOneOrangeDark}{someone}.''}
    {0.668}{0.292}{wrongly flagged}
  \fOnePanel{(1301,410)}{Orange}{Negation}{no vehicle; absence stated}{img_negation}
    {``The street appears\\to be empty, with\\\fOneKw{fOneOrangeDark}{no} other people or\\\fOneKw{fOneOrangeDark}{vehicles} visible\\in the image.''}
    {0.664}{0.332}{wrongly flagged}
\end{tikzpicture}}
\caption{\textbf{Hallucination detectors flag non-assertive mentions.}
Six illustrative held-out examples from LLaVA-1.5:
concrete, hallucinated, and four non-assertive subtypes
(hedged, depictive, purpose, and negation).
Bars show Lookback Lens scores before non-assertive training;
circles show scores after adding 20 non-assertive examples.
The dashed line marks the threshold calibrated at 10\%
concrete FPR. Training reduces the overall non-assertive
FPR from 40\% to 17\%.}
\label{fig:teaser}
\end{figure*}

Vision-language models (VLMs) are increasingly used for open-ended image description, and their outputs are evaluated by extracting the objects a caption names and matching them against ground-truth object annotations. This procedure, introduced by CHAIR~\citep{rohrbach2018object}, underlies
caption-level hallucination metrics~\citep{rohrbach2018object,kaul2024throne}, and the same labels are used to train and evaluate hallucination
detectors~\citep{zhang2025dhcp,fieback2024metatoken,zohrabi2026haloprobe}.
It rests on an implicit assumption: every object mention asserts that the object is present in the image.


Captions sometimes violate this assumption. A model may name an object without claiming that it is there: in ``a bridge for people to cross,'' the word ``people'' describes what the bridge is for, not who is in the image. We call such mentions \emph{non-assertive}, following work on factuality and veridicality in text~\citep{sauri2009factbank,demarneffe2012happen}, and distinguish four subtypes as shown in Figure \ref{fig:teaser}: (1) \emph{depictive}, where the object appears only as a representation (``a painting of a ship''); (2) \emph{purpose}, where the object is what something in the scene is for or is waiting for (``a bench for people to sit on''); (3) \emph{negation}, where the caption names the object to deny it (``no cars are visible''); and (4) \emph{hedged}, where the model marks its own uncertainty (``possibly a remote control''). Each mention thus has two independent properties: its \emph{usage}, existential (asserting that the object is present) or non-assertive, which depends only on the text, and its \emph{presence}, whether the object appears in the image. A hallucination detector should flag existential mentions of absent objects and nothing else.
Existing benchmarks leave this problem unaddressed for two reasons. First, hallucination datasets annotate object presence but not usage~\citep{li2023pope,wang2023amber}, so they cannot separate a non-assertive mention of an absent object from a genuine hallucination, and detectors trained on them inherit the conflation. Second, the metrics built on these labels penalise non-assertive language: CHAIR, and even THRONE~\citep{kaul2024throne}, which asks an LLM whether the caption implies each object, count about half of all non-assertive mentions in NAME-Bench as hallucinations (\S\ref{sec:res-eval}). As a result, a mitigation method could improve these metrics simply by making a model hedge or negate less often, without changing how often it hallucinates.

To address this gap, we introduce \textbf{NAME-Bench}
(Non-Assertive Mention Evaluation Benchmark), comprising
25,497 object mentions from captions generated by five
VLMs. Each mention is annotated for usage and presence,
with non-assertive mentions further classified by subtype.
We evaluate twelve hallucination detectors under a common
protocol and study whether their errors can be corrected
by training with non-assertive examples. The data is released.\footnote{\url{https://github.com/lab-flair/NAME_Bench}} Because existing benchmarks label mentions by presence alone, their hallucination rates and detection scores conflate non-assertive mentions with hallucinations. On \bench{}, every detector we test flags non-assertive mentions as hallucinations: trained detectors flag 56\% of them on average, and training-free detectors 31\%, against 10\% of concrete mentions.

Mixing \fm{} data does not fully fix the gap. The most direct remedy is to train detectors on mentions as they naturally occur, so that non-assertive mentions appear among the negatives at the same share as in the model's own captions. On LLaVA-1.5 this works: its natural share is only 20 examples, yet it halves the detectors' false-positive rate on non-assertive mentions, and about 100 examples bring it to 4--6\%. On the other four models, the natural share still leaves 16--46\% of non-assertive mentions flagged, and bringing it to 5--10\%, close to LLaVA-1.5's 4--6\%, takes five to ten times as much non-assertive data as LLaVA-1.5 needs, at a cost in recall
(\S\ref{sec:res-mixing}). A text filter that removes non-assertive mentions before scoring also removes real hallucinations, likely because these models word the two alike (\S\ref{sec:res-filter}). Hedged mentions remain the hardest case under both remedies. Our analysis suggests that these models describe absent objects in a recognisable way even when they do not
assert them, and that detectors pick up this signal (\S\ref{sec:res-findings}).

Our contributions are threefold. First, we define non-assertive object mentions and show that they are a distinct and systematic source of false positives for both hallucination detectors and hallucination metrics. Second, we release NAME-Bench and its evaluation protocol.
Third, we show that the cost of fixing this blind spot depends mainly on the captioning model rather than the detector. On LLaVA-1.5, non-assertive examples equal to 10\% of the training hallucinations bring the share of non-assertive mentions that trained detectors flag from 18--52\% to 4--6\%, with little change in how many hallucinations they catch. On the other four models, reaching a comparable rate (5–10\%) takes five to ten times more data than on LLaVA-1.5, and the detectors then catch fewer real hallucinations. 

\section{Related Work}

\paragraph{Hallucination Benchmarks and Metrics.}
Object hallucination is evaluated through free-form captions or direct object-existence questions. CHAIR~\citep{rohrbach2018object} matches captioned objects against image annotations, with extensions to open vocabularies~\citep{benkish2023openchair} and longer responses~\citep{kaul2024throne}. Polling benchmarks test object presence through binary, negative, or multi-object questions~\citep{li2023pope,lovenia2023nope,chen2024rope}. Other benchmarks assess attributes, relations, events, and open-ended responses~\citep{wang2023amber,jiang2024haleval,wu2024rbench,sun2023mmhal,liu2024lrv,jing2023faithscore}, or distinguish language-prior failures from perceptual errors~\citep{guan2024hallusionbench,liu2025phd}. Unlike polling, caption-based metrics must determine what a caption asserts, yet typically label mentions by object presence alone. Even THRONE~\citep{kaul2024throne}, which judges whether a caption implies an object's existence, counts over half of hedged and depictive mentions as hallucinations (Table~\ref{tab:chair-throne}).

\paragraph{Hallucination Detection.}
Detectors use VLM-internal signals, including attention to context or generated tokens~\citep{chuang2024lookback,hoang2026pas}, cross-modal and visual attention~\citep{zhang2025dhcp,jiang2025devils,prabhakaran2025vade}, token statistics and hidden representations~\citep{fieback2024metatoken,zohrabi2026haloprobe}, visual grounding~\citep{jiang2025interpreting}, and embedding similarity~\citep{park2026glsim}. HALP instead predicts hallucination risk before generation~\citep{kogilathota2026halp}. Existing training and evaluation commonly use presence-based labels~\citep{zhang2025dhcp,fieback2024metatoken,zohrabi2026haloprobe,gunjal2024mhaldetect}, treating \fm{} mentions like ordinary object claims. We examine whether detector scores reflect object absence even when the caption does not assert its presence.

\paragraph{Factuality and Non-assertion in Text.}
Factuality and veridicality research studies whether text presents events as certain, possible, or counterfactual~\citep{sauri2009factbank,sauri2012sure,demarneffe2012happen,rudinger2018factuality}. Related work addresses hedging and speculation~\citep{vincze2008bioscope,farkas2010conll}, negation scope~\citep{morante2009metalearning,morante2012sem}, and intensional verbs whose objects need not exist~\citep{zimmermann1993opacity}. Claim-level evaluation similarly verifies atomic assertions rather than all mentions~\citep{min2023factscore}. We apply these distinctions to object mentions in VLM captions to distinguish asserted presence from non-assertive usage.

\section{Benchmark}
\label{sec:data}
 

\bench{} tests whether detectors distinguish asserted presence from mere mention. It contains usage- and presence-labelled mentions from five VLMs and a common evaluation protocol.

\subsection{Task Definition}
\label{sec:def}

Given an image and a text prompt, a vision-language model generates a caption. An \emph{object mention} is an occurrence, in the caption, of a word that refers to one of a fixed set of object categories. A hallucination detector assigns each mention a score, from the caption and, for most detectors, the model's internal states, and flags the mention if the score exceeds a threshold. 

We describe each mention along two independent axes. Its \emph{usage} is determined by the caption text alone: a mention is \emph{existential} if the caption asserts that the object is in the image, and \emph{\fm{}} if the caption names the object without making that assertion. Its \emph{presence} is determined by the image alone: whether an instance of the object appears in it. Together the two axes give three classes: 1. \textbf{concrete}: existential and present; 2. \textbf{hallucinated}: existential and absent; 3. \textbf{\fm{}}: not asserted, whether or not the object is present.
A hallucination detector should flag the second class and nothing else. CHAIR~\citep{rohrbach2018object} labels
a mention by presence alone, so it cannot represent the third class: a \fm{} mention of an absent object is
counted as a hallucination. We keep \fm{} mentions of present objects as well, and record presence for every
mention, which lets us test whether a detector responds to the object's absence or to the caption's claim.

\paragraph{Subtypes.}
We distinguish four subtypes of \fm{} mention, according to why the caption names the object without
claiming that it is present (Figure~\ref{fig:teaser}). In a \textit{purpose} mention, the object is named
as the intended use or goal of something in the scene, including objects of intensional verbs
\citep{zimmermann1993opacity} (``ready to be enjoyed by someone'', ``people waiting
for the bus''). In a \textit{depictive} mention, the object appears only as a representation such as a
picture, a logo or written text (``a cake adorned with a picture of a dog'', ``a poster of a horse''). Here
the object's appearance may be visible in the image even though no instance of the object is present. In a
\textit{hedged} mention, the caption offers the object as uncertain, as in speculation
\citep{vincze2008bioscope,farkas2010conll} (``possibly watching birds'', ``there
may be a cat on the sofa''). In a \textit{negation} mention, the caption
names the object in order to deny it, placing it within the scope of a
negation \citep{morante2012sem} (``no people are visible'').

\subsection{Captioning and Mention Extraction}
\label{sec:data:captions}


\begin{table}[t]
\centering
\small
\setlength{\tabcolsep}{2pt}
\renewcommand{\arraystretch}{1.08}
\begin{tabular}{@{}lcc@{}}
\toprule
Model & \Fm{} share & Halluc.\ rate \\
\midrule
LLaVA-1.5-7B
    & 1.3 {\scriptsize[1.0, 1.6]}
    & 28.2 {\scriptsize[26.9, 29.3]} \\
Qwen2.5-VL-7B
    & 5.0 {\scriptsize[4.4, 5.8]}
    & 6.6 {\scriptsize[5.7, 7.5]} \\
Qwen2.5-VL-3B
    & 4.8 {\scriptsize[4.1, 5.6]}
    & 8.9 {\scriptsize[8.0, 9.9]} \\
InternVL3-8B
    & 4.2 {\scriptsize[3.6, 4.9]}
    & 7.0 {\scriptsize[6.2, 7.8]} \\
LLaVA-OneVision-2-8B
    & 4.4 {\scriptsize[3.7, 5.1]}
    & 5.7 {\scriptsize[4.9, 6.5]} \\
\bottomrule
\end{tabular}
\caption{Natural rates on captions of random images (2,000 per model), with 95\% image-bootstrap confidence intervals.
\textit{Non-assertive share}: fraction of object mentions that are
non-assertive. \textit{Halluc.\ rate}: fraction
of existential mentions that are hallucinated.}
\label{tab:natural_rates}
\end{table}

We build \bench{} from the captions of five models: LLaVA-1.5-7B~\cite{liu2024llava15}, Qwen2.5-VL-3B and
-7B~\cite{bai2025qwen25vl}, InternVL3-8B~\cite{zhu2025internvl3} and
LLaVA-OneVision-2-8B~\cite{an2026llavaonevision2nextgenerationperceptualintelligence}. 
The data are
\emph{on-policy}: each model's mentions come from captions that the model generated itself, not from
human-written captions or from another model's output.

Each model captions COCO~2017 images~\cite{lin2014coco} with the prompt ``Provide a detailed description of
this image.'', using greedy decoding and at most 128 new tokens. 
We find candidate mentions with a regular expression over the 80 COCO categories, using a curated list of synonyms and plurals (Appendix \ref{app:vocab}). 
Most of our forms also appear in CHAIR's synonym list \cite{rohrbach2018object}, but ours is smaller: it omits broad mappings that often do not refer to the category (e.g., \emph{table}, \emph{seat}, \emph{container}) as well as most fine-grained kinds and role words (e.g., dog breeds, \emph{chef}, \emph{passenger}).
The COCO instance annotations serve as ground truth for presence.

\subsection{Labelling Usage}
\label{sec:data:usage}

Usage is a property of the text, so it is judged from the caption alone, without the image, by
GPT-4o~\cite{openai2024gpt4o} at temperature 0 with six captions per call. 
The judge first decides whether the word refers to the object at all, which rules out,
for example, ``orange'' used as a colour. 
It then labels the usage as \textsc{existential} (presence is
asserted), \textsc{non-assertive} or \textsc{ambiguous}.
Finally, it returns the phrase that triggered its decision. Ambiguous mentions are discarded. Presence is taken from the COCO instance annotations of the image, following
POPE~\cite{li2023pope} and THRONE~\cite{kaul2024throne}: an object counts as present if its category is
annotated in the image. Combining
usage with presence gives the three classes of \S\ref{sec:def}: existential and present mentions are
concrete, existential and absent mentions are hallucinated, and \fm{} mentions are kept regardless of
presence.

\paragraph{Subtype Assignment.}
GPT-4o then assigns each \fm{} mention one of the four subtypes of \S\ref{sec:def}. The full prompt and the
mapping from its output to the subtypes are given in Appendix~\ref{app:prompt}. Two special cases are
removed. The first is compound expressions in which the object word does not refer to the object
(``hot dog'', ``orange juice'', ``teddy bear''), which GPT-4o-mini screens out. The second is \fm{}
mentions that fit none of the four subtypes, such as generic statements. Together these account for about 45\% of candidates initially labelled non-assertive, so \bench{} contains only \fm{} mentions
that fall clearly into one subtype. Excluded valid \fm{} mentions show similar detector failures (Appendix~\ref{app:other}).

\paragraph{Label Quality.}
To validate the labels, two annotators independently re-annotated 250 mentions sampled across the five
models (up to 10 per \fm{} category per model, plus 25 concrete and 25 hallucinated controls). Inter-annotator
agreement was substantial for the usage label (concrete, hallucinated or \fm{}; Cohen's $\kappa = 0.79$,
91.8\% raw agreement) and high for the \fm{} category ($\kappa = 0.90$, 92.4\%). Averaged over the two
annotators, the manual labels matched GPT-4o's usage label for 90.9\% of mentions and, where the usage label
was agreed, its category for 92.1\%. Category agreement ranged from 82.6\% for hedging to 98.9\% for negation.

\begin{table*}[t]
\centering
\small
\setlength{\tabcolsep}{4pt}
\begin{tabular}{l rrrr r rr}
\toprule
 & \multicolumn{5}{c}{\Fm{}} & & \\
\cmidrule(lr){2-6}
Model & Hedged & Depictive & Purpose & Negation & Total & Halluc. & Concrete \\
\midrule
LLaVA-1.5-7B         & 564 {\scriptsize(35.4)} & 265 {\scriptsize(16.6)} & 630 {\scriptsize(39.5)} & 135 {\scriptsize(8.5)} & 1{,}594 & 1{,}974 & 1{,}975 \\
Qwen2.5-VL-7B        & 958 {\scriptsize(51.3)} & 642 {\scriptsize(34.4)} & 182 {\scriptsize(9.8)} & 84 {\scriptsize(4.5)} & 1{,}866 & 1{,}983 & 1{,}878 \\
Qwen2.5-VL-3B        & 802 {\scriptsize(46.0)} & 516 {\scriptsize(29.6)} & 211 {\scriptsize(12.1)} & 216 {\scriptsize(12.4)} & 1{,}745 & 1{,}979 & 1{,}891 \\
InternVL3-8B         & 198 {\scriptsize(29.2)} & 307 {\scriptsize(45.3)} & 43 {\scriptsize(6.3)} & 130 {\scriptsize(19.2)} & 678 & 1{,}321 & 1{,}947 \\
LLaVA-OneVision-2-8B & 561 {\scriptsize(42.7)} & 450 {\scriptsize(34.2)} & 172 {\scriptsize(13.1)} & 131 {\scriptsize(10.0)} & 1{,}314 & 1{,}433 & 1{,}919 \\
\bottomrule
\end{tabular}
\caption{Mentions per model in \bench{}. Each \fm{} count is followed by its share of that model's \fm{} total (\%).}
\label{tab:dataset_counts}
\end{table*}

\begin{table*}[t]
\centering
\small
\setlength{\tabcolsep}{5pt}
\renewcommand{\arraystretch}{1.10}
\begin{tabular}{@{}lcc cc cc cc cc cc@{}}
\toprule
& \multicolumn{2}{c}{Halluc.\ Recall $\uparrow$}
& \multicolumn{2}{c}{\Fm{} (All) $\downarrow$}
& \multicolumn{2}{c}{Hedged $\downarrow$}
& \multicolumn{2}{c}{Depictive $\downarrow$}
& \multicolumn{2}{c}{Purpose $\downarrow$}
& \multicolumn{2}{c}{Negation $\downarrow$} \\
\cmidrule(lr){2-3}
\cmidrule(lr){4-5}
\cmidrule(lr){6-7}
\cmidrule(lr){8-9}
\cmidrule(lr){10-11}
\cmidrule(l){12-13}
Model
& C & T & C & T & C & T & C & T & C & T & C & T \\
\midrule
LLaVA-1.5-7B
& 92 & 100 & 46 & 55 & 59 & 59 & 36 & 59 & 37 & 62 & 53 & 1 \\
Qwen2.5-VL-7B
& 68 & 97 & 49 & 53 & 52 & 55 & 51 & 59 & 34 & 43 & 30 & 14 \\
Qwen2.5-VL-3B
& 74 & 98 & 51 & 49 & 54 & 57 & 51 & 60 & 38 & 39 & 53 & 2 \\
InternVL3-8B
& 69 & 98 & 54 & 52 & 51 & 62 & 52 & 66 & 36 & 56 & 68 & 0 \\
LLaVA-OneVision-2-8B
& 64 & 98 & 46 & 48 & 49 & 56 & 46 & 58 & 36 & 29 & 49 & 1 \\
\bottomrule
\end{tabular}

\caption{Columns: hallucination recall ($\uparrow$), then non-assertive mentions overall and per subtype ($\downarrow$). C: CHAIR; T: THRONE (majority vote).}
\label{tab:chair-throne}
\end{table*}


\subsection{Statistics}
\label{sec:data:stats}

\paragraph{Natural Rates.} Table~\ref{tab:natural_rates} reports how often each model produces \fm{} mentions on random images.
They make up 1.3\% of LLaVA-1.5's object mentions and 4.2--5.0\% of the four newer models', while the
hallucination rate falls from 28.2\% for LLaVA-1.5 to 5.7--8.9\% for the newer models. On the four newer models, non-assertive mentions are thus far more common relative to the hallucinations a detector is meant to find than on LLaVA-1.5.

\paragraph{Evaluation Sets.}
Because non-assertive mentions are rare, we construct
evaluation sets that oversample them to enable
subtype-level analysis. Across five models, NAME-Bench
contains 25,497 mentions, including 678--1,866
non-assertive mentions per model (Table~\ref{tab:dataset_counts}).
The subtype distribution varies across models,
with hedged and depictive mentions generally prevalent. The distribution per concept is in Appendix \ref{app:concepts}.

\subsection{Evaluation Protocol}
\label{sec:data:protocol}

Every detector assigns each mention a score. We split each model's mentions into training (53.3\%), calibration
(13.3\%) and held-out test (33.3\%) sets. For each detector and model we set the decision threshold on the
calibration set so that 10\% of its concrete mentions are flagged, which places all detectors at the same
false-alarm rate on ordinary grounded mentions. At this threshold we report, on the held-out test set, the
\emph{\fm{} FPR}, the percentage of \fm{} mentions flagged as hallucinated (lower is better), pooled and per
category, and the \emph{hallucination TPR}, the percentage of hallucinated mentions flagged (higher is better).
Results are means over 3 seeds.
Hallucinations tend to occur later in a caption than concrete mentions, so token position alone carries
signal~\cite{zohrabi2026haloprobe}. We therefore repeat key comparisons on a \emph{position-matched} subset,
in which the classes being compared have the same distribution of token positions (Appendix \ref{app:position}).

\section{Re-Benchmarking Hallucination Metrics and Detectors}
\label{sec:results}
\providecommand{\rtodo}[1]{\textcolor{red}{[TODO: #1]}}

We now use \bench{} to re-evaluate how existing metrics and detectors treat \fm{} mentions: two
standard metrics, CHAIR and THRONE, and the twelve detectors described below.

\paragraph{Detectors.}
\begin{table*}[t]
\centering\small\setlength{\tabcolsep}{2.5pt}
\resizebox{\textwidth}{!}{%
\begin{tabular}{l|ccccc|ccccc|ccccc|ccccc|ccccc}
\toprule
 & \multicolumn{5}{c|}{LLaVA-1.5-7B} & \multicolumn{5}{c|}{Qwen2.5-VL-7B} & \multicolumn{5}{c|}{Qwen2.5-VL-3B} & \multicolumn{5}{c|}{InternVL3-8B} & \multicolumn{5}{c}{LLaVA-OneVision-2} \\
Detector & H & D & P & N & All & H & D & P & N & All & H & D & P & N & All & H & D & P & N & All & H & D & P & N & All \\
\midrule
Lookback Lens       & 64 & 27 & 24 & 45 & \textbf{40} & 71 & 75 & 39 & 29 & \textbf{67} & 81 & 57 & 59 & 37 & \textbf{66} & 80 & 68 & 67 & 11 & \textbf{62} & 88 & 77 & 76 & 58 & \textbf{80} \\
Hidden state (mid)  & 64 & 43 & 46 & 48 & \textbf{52} & 63 & 66 & 38 & 28 & \textbf{60} & 64 & 51 & 42 & 14 & \textbf{51} & 70 & 72 & 52 & 22 & \textbf{61} & 72 & 72 & 40 & 36 & \textbf{63} \\
Hidden state (last) & 49 & 31 & 27 & 3  & \textbf{34} & 69 & 68 & 48 & 32 & \textbf{65} & 70 & 57 & 43 & 14 & \textbf{56} & 69 & 61 & 56 & 1  & \textbf{53} & 76 & 76 & 53 & 24 & \textbf{68} \\
DHCP                & 28 & 14 & 10 & 14 & \textbf{18} & 67 & 53 & 45 & 49 & \textbf{59} & 72 & 41 & 62 & 71 & \textbf{62} & 75 & 57 & 56 & 42 & \textbf{60} & 84 & 55 & 68 & 66 & \textbf{71} \\
HALP                & 18 & 38 & 15 & 33 & \textbf{22} & 15 & 13 & 12 & 7  & \textbf{13} & 12 & 12 & 6  & 10 & \textbf{11} & 24 & 17 & 20 & 9  & \textbf{18} & 17 & 18 & 10 & 15 & \textbf{16} \\
Text-only LM        & 56 & 33 & 42 & 52 & \textbf{46} & 56 & 66 & 33 & 19 & \textbf{56} & 59 & 50 & 42 & 11 & \textbf{48} & 65 & 65 & 44 & 10 & \textbf{54} & 64 & 76 & 40 & 20 & \textbf{60} \\
\midrule
SVAR                & 48 & 4  & 81 & 97 & \textbf{57} & 67 & 18 & 74 & 99 & \textbf{53} & 59 & 18 & 66 & 98 & \textbf{53} & 58 & 14 & 83 & 100 & \textbf{46} & 66 & 22 & 74 & 100 & \textbf{57} \\
VADE                & 38 & 2  & 72 & 84 & \textbf{48} & 56 & 14 & 68 & 91 & \textbf{45} & 52 & 15 & 60 & 89 & \textbf{47} & 47 & 15 & 48 & 64 & \textbf{35} & 46 & 11 & 51 & 74 & \textbf{38} \\
Internal Conf.      & 49 & 23 & 49 & 84 & \textbf{47} & 23 & 33 & 29 & 78 & \textbf{30} & 28 & 28 & 37 & 74 & \textbf{35} & 25 & 19 & 13 & 39 & \textbf{24} & 34 & 38 & 36 & 56 & \textbf{37} \\
GLSim               & 20 & 2  & 17 & 63 & \textbf{19} & 16 & 28 & 24 & 31 & \textbf{21} & 23 & 31 & 4  & 4  & \textbf{21} & 20 & 32 & 6  & 84 & \textbf{36} & 27 & 47 & 39 & 72 & \textbf{39} \\
PAS                 & 17 & 2  & 20 & 1  & \textbf{14} & 24 & 19 & 20 & 58 & \textbf{24} & 20 & 18 & 28 & 31 & \textbf{22} & 21 & 18 & 10 & 85 & \textbf{30} & 15 & 11 & 23 & 29 & \textbf{16} \\
NLL                 & 23 & 26 & 5  & 10 & \textbf{16} & 34 & 27 & 19 & 16 & \textbf{29} & 23 & 23 & 15 & 13 & \textbf{21} & 21 & 19 & 12 & 11 & \textbf{18} & 22 & 20 & 5  & 7  & \textbf{17} \\
Entropy             & 32 & 34 & 3  & 6  & \textbf{19} & 33 & 33 & 12 & 4  & \textbf{30} & 23 & 32 & 7  & 6  & \textbf{21} & 18 & 21 & 8  & 4  & \textbf{16} & 24 & 29 & 3  & 7  & \textbf{21} \\
\bottomrule
\end{tabular}}
\caption{\Fm{} FPR (\%, $\downarrow$) per category for detectors built without \fm{} training data. Top: trained probes; bottom:
training-free detectors. H: hedged, D: depictive, P: purpose, N: negation; All: all \fm{} mentions of the model
(weighted by category size). HALP scores the image before the caption is generated and so gives every mention in a
caption the same score.}
\label{tab:main}
\end{table*}

We evaluate twelve detectors in two groups. \emph{Trained probes} fit a classifier on features of the
captioning model at the mention: Lookback Lens~\cite{chuang2024lookback} on attention ratios, hidden-state
probes at a middle and at the last layer, DHCP~\cite{zhang2025dhcp} on cross-modal attention, and a probe on
the text-only language model, which never sees the image. \emph{Training-free} detectors compute a score
with no fitted parameters: SVAR~\cite{jiang2025devils}, VADE~\cite{prabhakaran2025vade} and Internal
Confidence~\cite{jiang2025interpreting} measure visual grounding, GLSim~\cite{park2026glsim} embedding
similarity, PAS~\cite{hoang2026pas} attention to previously generated tokens, and NLL and entropy
token-level uncertainty. We also report HALP~\cite{kogilathota2026halp} for reference. HALP scores the
prompt before any caption is generated, so it produces one score per image and cannot tell the mentions of
a caption apart. More details in Appendix \ref{app:detectors}.

\subsection{Results on Existing Metrics and Detectors}
\label{sec:res-eval}

\paragraph{Effect on CHAIR and THRONE.}
The conflation of mention and claim is built into the standard metrics (Table~\ref{tab:chair-throne}). CHAIR and
THRONE each count about half of all \fm{} mentions as hallucinations (CHAIR 46--54\%, THRONE 48--55\%), while neither counts any concrete mention, by construction, since concrete objects are annotated as present. The two fail in different places. THRONE resolves explicit negation, which CHAIR
counts in 30--68\% of cases (THRONE 0--14\%), but it reads depictions, and in most models purposes, as claims more
often than CHAIR does (depictive 58--66\% vs.\ 36--52\%). Neither handles hedges (49--62\%). An LLM judge thus
removes the easiest case, but a metric that asks whether the text names or implies an object, rather than whether it
asserts it, will keep counting \fm{} mentions as hallucinations. THRONE's
default unanimous vote leaves 55--71\% of hedged, depictive and purpose
mentions unscored.
Following POPE and THRONE, we treat an object as present if it appears in the COCO instance annotations; CHAIR additionally counts objects named in the reference captions, which is why it recovers only 64--92\% of our hallucinations. Combining the CHAIR rates in Table~\ref{tab:chair-throne} with the natural rates in Table~\ref{tab:natural_rates}, we find that \fm{} mentions make up 28--37\% of the objects CHAIR counts as hallucinated on the four newer models, against about 2\% on LLaVA-1.5. Additional THRONE analyses are in Appendix~\ref{app:throne}.

\paragraph{Every detector mistakes \fm{} mentions for hallucinations.}

We first evaluate the detectors as published, that is, trained or built without any \fm{} examples
(Table~\ref{tab:main}). On all five models, every detector flags a substantial share of \fm{} mentions as
hallucinated. Which mentions are flagged depends on the kind of detector. Apart from HALP, which cannot tell the
mentions of a caption apart, trained probes fail most on hedged and depictive mentions, flagging 56--88\% and
41--77\% of them on the four newer models. The grounding-based detectors SVAR and VADE show the reverse pattern:
SVAR flags almost every negation (97--100\%) and both flag 48--83\% of purpose mentions, but they flag few
depictive mentions (2--22\%). NLL and entropy are the only detectors that stay below 35\% in every category.

\begin{table*}[t]
\centering\small\setlength{\tabcolsep}{2.5pt}
\resizebox{\textwidth}{!}{%
\begin{tabular}{l|ccccc|ccccc|ccccc|ccccc|ccccc}
\toprule
 & \multicolumn{5}{c|}{LLaVA-1.5-7B} & \multicolumn{5}{c|}{Qwen2.5-VL-7B} & \multicolumn{5}{c|}{Qwen2.5-VL-3B} & \multicolumn{5}{c|}{InternVL3-8B} & \multicolumn{5}{c}{LLaVA-OneVision-2} \\
Detector & 0 & nat. & 10 & 30 & max & 0 & nat. & 10 & 30 & max & 0 & nat. & 10 & 30 & max & 0 & nat. & 10 & 30 & max & 0 & nat. & 10 & 30 & max \\
\midrule
\multicolumn{26}{l}{\textit{\Fm{} FPR (\%) $\downarrow$}} \\
\midrule
Lookback Lens       & 40 & 17 & 4 & 2 & 1 & 67 & 32 & 31 & 15 & 7 & 66 & 31 & 31 & 15 & 5 & 62 & 37 & 38 & 15 & 9 & 80 & 38 & 39 & 17 & 7 \\
Hidden state (mid)  & 52 & 23 & 6 & 3 & 2 & 60 & 23 & 24 & 14 & 7 & 51 & 23 & 22 & 13 & 7 & 61 & 19 & 19 & 13 & 10 & 63 & 28 & 29 & 18 & 9 \\
Hidden state (last) & 34 & 17 & 5 & 3 & 2 & 65 & 24 & 24 & 14 & 7 & 56 & 23 & 23 & 15 & 7 & 53 & 20 & 21 & 12 & 9 & 68 & 30 & 31 & 17 & 9 \\
DHCP                & 18 & 11 & 5 & 2 & 1 & 59 & 34 & 34 & 15 & 7 & 62 & 28 & 27 & 13 & 6 & 60 & 23 & 22 & 10 & 6 & 71 & 46 & 46 & 22 & 8 \\
Text-only LM        & 46 & 18 & 5 & 2 & 1 & 56 & 17 & 18 & 9 & 5 & 48 & 22 & 22 & 14 & 7 & 54 & 16 & 18 & 10 & 8 & 60 & 26 & 27 & 14 & 7 \\
\cmidrule(lr){1-26}
HALP                & 22 & 15 & 12 & 8 & 7 & 13 & 13 & 12 & 12 & 9 & 11 & 12 & 10 & 10 & 9 & 18 & 17 & 19 & 13 & 16 & 16 & 13 & 14 & 11 & 9 \\
\midrule
\multicolumn{26}{l}{\textit{Hallucination TPR (\%) $\uparrow$}} \\
\midrule
Lookback Lens       & 54 & 54 & 52 & 50 & 49 & 64 & 58 & 58 & 56 & 51 & 64 & 64 & 64 & 59 & 52 & 55 & 54 & 54 & 49 & 47 & 71 & 68 & 67 & 64 & 61 \\
Hidden state (mid)  & 49 & 50 & 50 & 51 & 52 & 69 & 65 & 65 & 64 & 57 & 64 & 61 & 61 & 57 & 50 & 63 & 60 & 59 & 60 & 60 & 75 & 71 & 71 & 68 & 64 \\
Hidden state (last) & 46 & 43 & 42 & 43 & 43 & 72 & 66 & 65 & 65 & 57 & 64 & 60 & 61 & 58 & 54 & 62 & 59 & 59 & 57 & 57 & 77 & 70 & 70 & 67 & 61 \\
DHCP                & 43 & 37 & 40 & 42 & 43 & 59 & 60 & 60 & 52 & 45 & 56 & 56 & 55 & 49 & 46 & 53 & 47 & 45 & 46 & 38 & 62 & 58 & 59 & 54 & 46 \\
Text-only LM        & 35 & 35 & 36 & 35 & 36 & 54 & 52 & 52 & 49 & 46 & 54 & 50 & 50 & 47 & 44 & 50 & 48 & 49 & 50 & 49 & 68 & 62 & 63 & 60 & 56 \\
\cmidrule(lr){1-26}
HALP                & 14 & 12 & 12 & 10 & 12 & 21 & 22 & 19 & 19 & 21 & 15 & 17 & 15 & 18 & 18 & 19 & 19 & 22 & 19 & 20 & 23 & 24 & 22 & 20 & 21 \\
\bottomrule
\end{tabular}}
\caption{\Fm{} FPR (top) and hallucination TPR (bottom) of the trained detectors as \fm{} mentions are added to the
training data as negatives. Columns give the number of \fm{} training examples relative to the training
hallucinations (\%). \emph{nat.}: natural share (20 examples for LLaVA-1.5, 76--108 for the other models).
\emph{max}: all available \fm{} examples, for models with fewer \fm{} than hallucination examples.
}
\label{tab:mixing-side}
\end{table*}
\paragraph{Several detectors mostly measure caption position.} A mention's token position alone separates concrete
from hallucinated mentions (AUC 0.81 on LLaVA-1.5 and 0.67--0.72 on the newer models), because hallucinations occur
late in long captions. On a position-matched subset, PAS falls to chance in every model (0.51--0.53), as do SVAR
(0.51) and VADE (0.48) on LLaVA-1.5, whereas the trained probes keep most of their signal (hidden-state probe
0.76--0.89; Table~\ref{tab:position}). This confirms the positional confound reported by \citet{zohrabi2026haloprobe}. Non-assertive mentions sit late in captions, like hallucinations (median relative position 0.46--0.55 vs.\ 0.18--0.22 for concrete mentions), but position does not explain their FPR: on a subset in which concrete, hallucinated and \fm{} mentions are matched for position, the trained probes' \fm{} FPR changes by at most 3 points (48--79\% on the newer models), while concrete mentions at the same positions are flagged at 11--21\%.

\section{Correcting the Detectors}
\label{sec:correct}
\begin{table}[t]
\centering\small\setlength{\tabcolsep}{4pt}
\resizebox{\columnwidth}{!}{%
\begin{tabular}{l cccc}
\toprule
Model & Hedged & Depictive & Purpose & Negation \\
\midrule
LLaVA-1.5-7B         & 3 $\rightarrow$ 19  & 4 $\rightarrow$ 18  & 1 $\rightarrow$ 4   & 0 $\rightarrow$ 2 \\
Qwen2.5-VL-7B        & 18 $\rightarrow$ 44 & 10 $\rightarrow$ 35 & 8 $\rightarrow$ 10  & 4 $\rightarrow$ 7 \\
Qwen2.5-VL-3B        & 22 $\rightarrow$ 47 & 8 $\rightarrow$ 26  & 11 $\rightarrow$ 13 & 1 $\rightarrow$ 5 \\
InternVL3-8B         & 19 $\rightarrow$ 56 & 9 $\rightarrow$ 39  & 32 $\rightarrow$ 41 & 0 $\rightarrow$ 3 \\
LLaVA-OneVision-2-8B & 24 $\rightarrow$ 58 & 15 $\rightarrow$ 39 & 15 $\rightarrow$ 19 & 2 $\rightarrow$ 8 \\
\bottomrule
\end{tabular}}
\caption{Non-assertive FPR (\%, $\downarrow$) on each subtype when it is included in (seen) → excluded from (unseen) training. Both conditions use a 30\% budget; in the unseen condition, the same total number of non-assertive examples is drawn from the other three subtypes. Mean over five trained probes.}
\label{tab:loco}
\end{table}

Usage could be judged by an external LLM, as in our labelling pipeline (\S\ref{sec:data}), but detectors are typically used during decoding, where a model call per mention is costly. We therefore first ask whether the detectors can be corrected directly, and then compare against a text filter placed in front of them.

\paragraph{Training Budget.}
Trained probes are first fitted on concrete and hallucinated mentions only, as published. To measure how
much \fm{} data is needed to correct them, we then add \fm{} mentions to the training set as negatives. The
budget is expressed as the number of \fm{} training examples relative to the number of training
hallucinations: 0\%, 10\%, 30\% and 100\% (or the maximum available). We add a \emph{natural} budget, in
which \fm{} mentions make up the same share of the training mentions as they do in the model's own captions
(Table~\ref{tab:natural_rates}). This is 20 examples for LLaVA-1.5 and, for the newer models, about 10\% of
the training hallucinations. Training-free detectors have no training stage and are evaluated once.

\subsection{Training on Non-Assertive Examples}
\label{sec:res-mixing}

\paragraph{Mixing works on LLaVA-1.5 but needs five to ten times more
data on the other models (Table~\ref{tab:mixing-side}).} 
At its
natural share, only 20 examples, the non-assertive FPR of Lookback Lens
drops from 40\% to 17\% and that of the mid-layer hidden-state probe
from 52\% to 23\%; at 10\% of the hallucination count (about 100
examples), all five trained probes fall to 4--6\%. On the four other
models the same recipe is far less effective. The five probes start at
48--80\%, still flag 16--46\% at the natural share and 18--46\% at 10\%
mixing, and reach 5--10\% only at the largest budget (53--100\% of the
hallucination count), five to ten times the share LLaVA-1.5 needs. This
also costs recall: at maximum mixing, the hallucination TPR of Lookback
Lens falls from 64\% to 51\% on Qwen2.5-VL-7B and from 71\% to 61\% on
LLaVA-OneVision-2, against 54\% to 49\% on LLaVA-1.5, where the other
four probes change by at most 3 points.
A teacher-forced control using LLaVA-1.5 captions
substantially reduces this cross-model gap,
suggesting that caption distribution contributes
to the difference (Appendix \ref{app:teacher-forced}).
\begin{table}[t]
\centering
\small
\footnotesize
\setlength{\tabcolsep}{2pt}
\renewcommand{\arraystretch}{1.03}

\begin{tabular}{@{}llcccccc@{}}
\toprule
 & & \multicolumn{3}{c}{Hedged}
     & \multicolumn{3}{c}{All \fm{}} \\
\cmidrule(lr){3-5}\cmidrule(lr){6-8}
Probe & Model & Abs. & Pres. & AUC & Abs. & Pres. & AUC \\
\midrule
\multirow{5}{*}{\shortstack[l]{Hidden\\state (mid)}}
 & LLaVA-1.5 & 64 & 62 & 0.52 & 55 & 43 & 0.58 \\
 & Qwen-7B   & 74 & 45 & 0.71 & 68 & 43 & 0.68 \\
 & Qwen-3B   & 73 & 43 & 0.70 & 59 & 33 & 0.66 \\
 & InternVL3 & 80 & 44 & 0.78 & 64 & 54 & 0.60 \\
 & LLaVA-OV2 & 83 & 47 & 0.77 & 74 & 42 & 0.73 \\
\midrule
\multirow{5}{*}{\shortstack[l]{Text-only\\LM}}
 & LLaVA-1.5 & 55 & 60 & 0.50 & 48 & 43 & 0.53 \\
 & Qwen-7B   & 61 & 47 & 0.61 & 60 & 45 & 0.61 \\
 & Qwen-3B   & 65 & 45 & 0.63 & 52 & 39 & 0.57 \\
 & InternVL3 & 76 & 38 & 0.75 & 59 & 40 & 0.62 \\
 & LLaVA-OV2 & 73 & 45 & 0.68 & 67 & 46 & 0.65 \\
\bottomrule
\end{tabular}

\caption{Non-assertive mentions flagged (\%, $\downarrow$) when the object is absent (Abs.) or present (Pres.) in the image. Results for the mid-layer hidden-state and text-only LM probes, trained without non-assertive examples. AUC measures absent-vs.-present score separation.}

\label{tab:absent-present}
\end{table}
\paragraph{Hedged and depictive mentions need their own examples.}
Excluding negation or purpose from training increases FPR
by only 2--9 points, unlike hedged and depictive mentions (Table \ref{tab:loco}).
Their FPR rises two- to six-fold in every model: hedged from 3\% to 19\% on LLaVA-1.5 and from 18--24\% to
44--58\% on the newer models, depictive from 4\% to 18\% and from 8--15\% to 26--39\%. Training data should
therefore cover every category a model produces, above all hedged and depictive mentions.

\subsection{Text Filter}
\label{sec:res-filter}

\paragraph{A text filter helps training-free detectors, but removes hallucinations on newer models.}
Mixing cannot change training-free detectors. Because usage depends only on the text, a \emph{text filter} can be placed in front of any detector: a probe on the text-only language model, trained on the natural share of \fm{} examples; mentions it marks as \fm{} are never flagged (Table~\ref{tab:filter}, Appendix~\ref{app:filter}). The filter catches 62--69\% of \fm{} mentions, reducing training-free detector FPR from 14--57\% to 2--19\%. However, it also removes 0.2\% of hallucinations on LLaVA-1.5 and 2--4\% on the newer models. On the latter, the text-only probe outperforms position alone (AUC 0.82--0.88 vs.\ 0.67--0.72), unlike LLaVA-1.5 (both 0.81; Table~\ref{tab:position}). This suggests overlapping textual cues, especially for hedges (50--58\% filtered vs.\ 67\% on LLaVA-1.5).


%
\subsection{Why Detectors Fail}
\label{sec:res-findings}

\paragraph{Newer models word, and to a lesser extent encode, an object's absence even when it is only mentioned.} Among \fm{}
mentions, the hidden-state probe flags those whose object is absent from the image far more often than those whose
object is present on all four newer models (hedges: 73--83\% vs.\ 43--47\%, AUC 0.70--0.78), but not on LLaVA-1.5
(64\% vs.\ 62\%, AUC 0.52; Table~\ref{tab:absent-present}). 
A text-only probe also separates absent from present hedges
(AUC 0.61--0.75 versus 0.50 on LLaVA-1.5).
The multimodal state adds a modest increment on most models (+0.03 to +0.10 AUC on hedges). The wording and, to a lesser extent, the internal state of newer models
thus carry the same signal that marks a hallucination, which is consistent with mixing separating the two
only partly on their captions while it succeeds on LLaVA's.

\paragraph{The failure persists when hedges are treated as claims.}
Hedges resist mixing (8--14\% FPR versus 2--3\% on LLaVA-1.5; Table~\ref{tab:cat4-mixing-sub}, Appendix~\ref{app:cat-mixing}) and corrections transfer poorly from other subtypes (Table~\ref{tab:loco}). Since hedges such as ``possibly a
bird'' may also be interpreted as weak claims, we test the opposite labelling
convention, treating hedges as existential mentions and retraining the probes.
The failure persists: the five probes still flag 29--58\% of the remaining
non-assertive mentions without \fm{} training and 12--33\% at the natural
share on the four other models (Appendix~\ref{app:hedgeclaim}).
Thus, our conclusions hold under either convention.

\section{Conclusion}
\label{sec:conclusion}
Object hallucination benchmarks and detectors often treat every object mention as a claim that the object is present. \bench{} shows that this assumption fails: captions can name objects without asserting their presence, yet existing metrics and detectors frequently flag such mentions as hallucinations. On the newer models, they make up 28--37\% of CHAIR's hallucinations, and correcting the detectors requires five to ten times more non-assertive data than on LLaVA-1.5, at a cost in recall. Our analysis suggests that these mentions share textual and internal signals with hallucinations, so hallucination evaluation should distinguish what a caption asserts from what it merely mentions.


\section*{Limitations}
Our analysis is restricted to mentions of the 80 COCO object categories in captions of COCO images, generated by
five open models with a single prompt and at most 128 new tokens. Extending \bench{} to attributes, relations,
other image domains and closed models is a natural next step. Beyond this scope, our measurements rest on two
sources of annotation, each with its own limits. The usage labels come from GPT-4o, and we validate them by
having two annotators independently re-annotate 250 mentions, reaching 90.9\% agreement with GPT-4o on usage
(inter-annotator $\kappa = 0.79$). Object presence, in turn, is taken from
the COCO instance annotations, which miss some objects. In addition, about 45\% of \fm{} mentions are excluded
because they are compounds or fit no single subtype. Finally, whether a hedge counts as a claim is a matter of
convention: we label hedges as \fm{}, and we test the opposite convention on all five models in
Appendix~\ref{app:hedgeclaim}. We release \bench{} to support detectors that distinguish a model's
claims from its mentions.

\bibliography{custom}

\begin{thebibliography}{42}
\providecommand{\natexlab}[1]{#1}

\bibitem[{An et~al.(2026)An, Xie, Tang, Yan, Tan, Zhu, Chen, Zhao, Qin, Yang et~al.}]{an2026llavaonevision2nextgenerationperceptualintelligence}
Xiang An, Yin Xie, Feilong Tang, Yunyao Yan, Huajie Tan, Didi Zhu, Changrui Chen, Xiuwei Zhao, Bin Qin, Kaicheng Yang, et~al. 2026.
\newblock Llava-onevision-2: Towards next-generation perceptual intelligence.
\newblock \emph{arXiv preprint arXiv:2605.25979}.

\bibitem[{Bai et~al.(2025)Bai, Chen, Liu, Wang, Ge, Song, Dang, Wang, Wang, Tang et~al.}]{bai2025qwen25vl}
Shuai Bai, Keqin Chen, Xuejing Liu, Jialin Wang, Wenbin Ge, Sibo Song, Kai Dang, Peng Wang, Shijie Wang, Jun Tang, et~al. 2025.
\newblock Qwen2.5-vl technical report.
\newblock \emph{arXiv preprint arXiv:2502.13923}.

\bibitem[{Ben-Kish et~al.(2024)Ben-Kish, Yanuka, Alper, Giryes, and Averbuch-Elor}]{benkish2023openchair}
Assaf Ben-Kish, Moran Yanuka, Morris Alper, Raja Giryes, and Hadar Averbuch-Elor. 2024.
\newblock Mitigating open-vocabulary caption hallucinations.
\newblock In \emph{Proceedings of the 2024 Conference on Empirical Methods in Natural Language Processing}, pages 22680--22698.

\bibitem[{Chen et~al.(2024)Chen, Ma, Zhang, Xu, Yang, Fouhey, Chai, and Qian}]{chen2024rope}
Xuweiyi Chen, Ziqiao Ma, Xuejun Zhang, Sihan Xu, Jianing Yang, David~F Fouhey, Joyce Chai, and Shengyi Qian. 2024.
\newblock Multi-object hallucination in vision language models.
\newblock \emph{Advances in Neural Information Processing Systems}, 37:44393--44418.

\bibitem[{Chuang et~al.(2024)Chuang, Qiu, Hsieh, Krishna, Kim, and Glass}]{chuang2024lookback}
Yung-Sung Chuang, Linlu Qiu, Cheng-Yu Hsieh, Ranjay Krishna, Yoon Kim, and James Glass. 2024.
\newblock Lookback lens: Detecting and mitigating contextual hallucinations in large language models using only attention maps.
\newblock In \emph{Proceedings of the 2024 Conference on Empirical Methods in Natural Language Processing}, pages 1419--1436.

\bibitem[{De~Marneffe et~al.(2012)De~Marneffe, Manning, and Potts}]{demarneffe2012happen}
Marie-Catherine De~Marneffe, Christopher~D Manning, and Christopher Potts. 2012.
\newblock Did it happen? the pragmatic complexity of veridicality assessment.
\newblock \emph{Computational linguistics}, 38(2):301--333.

\bibitem[{Farkas et~al.(2010)Farkas, Vincze, M{\'o}ra, Csirik, and Szarvas}]{farkas2010conll}
Rich{\'a}rd Farkas, Veronika Vincze, Gy{\"o}rgy M{\'o}ra, J{\'a}nos Csirik, and Gy{\"o}rgy Szarvas. 2010.
\newblock The conll-2010 shared task: learning to detect hedges and their scope in natural language text.
\newblock In \emph{Proceedings of the fourteenth conference on computational natural language learning--Shared task}, pages 1--12.

\bibitem[{Fieback et~al.(2024)Fieback, Spiegelberg, and Gottschalk}]{fieback2024metatoken}
Laura Fieback, Jakob Spiegelberg, and Hanno Gottschalk. 2024.
\newblock Metatoken: Detecting hallucination in image descriptions by meta classification.
\newblock \emph{arXiv preprint arXiv:2405.19186}.

\bibitem[{Guan et~al.(2024)Guan, Liu, Wu, Xian, Li, Liu, Wang, Chen, Huang, Yacoob et~al.}]{guan2024hallusionbench}
Tianrui Guan, Fuxiao Liu, Xiyang Wu, Ruiqi Xian, Zongxia Li, Xiaoyu Liu, Xijun Wang, Lichang Chen, Furong Huang, Yaser Yacoob, et~al. 2024.
\newblock Hallusionbench: an advanced diagnostic suite for entangled language hallucination and visual illusion in large vision-language models.
\newblock In \emph{2024 IEEE/CVF Conference on Computer Vision and Pattern Recognition (CVPR)}, pages 14375--14385. IEEE.

\bibitem[{Gunjal et~al.(2024)Gunjal, Yin, and Bas}]{gunjal2024mhaldetect}
Anisha Gunjal, Jihan Yin, and Erhan Bas. 2024.
\newblock Detecting and preventing hallucinations in large vision language models.
\newblock In \emph{Proceedings of the AAAI conference on artificial intelligence}, volume~38, pages 18135--18143.

\bibitem[{Hoang et~al.(2026)Hoang, Vu, Thai, and Bhattarai}]{hoang2026pas}
Nhat Hoang, Minh Vu, My~T Thai, and Manish Bhattarai. 2026.
\newblock Pas: Prelim attention score for detecting object hallucinations in large vision-language models.
\newblock In \emph{Proceedings of the IEEE/CVF Conference on Computer Vision and Pattern Recognition}, pages 18273--18283.

\bibitem[{Hurst et~al.(2024)Hurst, Lerer, Goucher, Perelman, Ramesh, Clark, Ostrow, Welihinda, Hayes, Radford et~al.}]{openai2024gpt4o}
Aaron Hurst, Adam Lerer, Adam~P Goucher, Adam Perelman, Aditya Ramesh, Aidan Clark, AJ~Ostrow, Akila Welihinda, Alan Hayes, Alec Radford, et~al. 2024.
\newblock Gpt-4o system card.
\newblock \emph{arXiv preprint arXiv:2410.21276}.

\bibitem[{Jiang et~al.(2024)Jiang, Jia, Dong, Ye, Xu, Yan, Zhang, and Zhang}]{jiang2024haleval}
Chaoya Jiang, Hongrui Jia, Mengfan Dong, Wei Ye, Haiyang Xu, Ming Yan, Ji~Zhang, and Shikun Zhang. 2024.
\newblock Hal-eval: A universal and fine-grained hallucination evaluation framework for large vision language models.
\newblock In \emph{Proceedings of the 32nd ACM International Conference on Multimedia}, pages 525--534.

\bibitem[{Jiang et~al.(2025{\natexlab{a}})Jiang, Kachinthaya, Petryk, and Gandelsman}]{jiang2025interpreting}
Nick Jiang, Anish Kachinthaya, Suzanne Petryk, and Yossi Gandelsman. 2025{\natexlab{a}}.
\newblock Interpreting and editing vision-language representations to mitigate hallucinations.
\newblock In \emph{International Conference on Learning Representations}, volume 2025, pages 63582--63605.

\bibitem[{Jiang et~al.(2025{\natexlab{b}})Jiang, Chen, Zhu, Luo, Shen, and Yang}]{jiang2025devils}
Zhangqi Jiang, Junkai Chen, Beier Zhu, Tingjin Luo, Yankun Shen, and Xu~Yang. 2025{\natexlab{b}}.
\newblock Devils in middle layers of large vision-language models: Interpreting, detecting and mitigating object hallucinations via attention lens.
\newblock In \emph{2025 IEEE/CVF Conference on Computer Vision and Pattern Recognition (CVPR)}, pages 25004--25014. IEEE.

\bibitem[{Jing et~al.(2024)Jing, Li, Chen, and Du}]{jing2023faithscore}
Liqiang Jing, Ruosen Li, Yunmo Chen, and Xinya Du. 2024.
\newblock Faithscore: Fine-grained evaluations of hallucinations in large vision-language models.
\newblock In \emph{Findings of the Association for Computational Linguistics: EMNLP 2024}, pages 5042--5063.

\bibitem[{Kaul et~al.(2024)Kaul, Li, Yang, Dukler, Swaminathan, Taylor, and Soatto}]{kaul2024throne}
Prannay Kaul, Zhizhong Li, Hao Yang, Yonatan Dukler, Ashwin Swaminathan, Christopher~J Taylor, and Stefano Soatto. 2024.
\newblock Throne: An object-based hallucination benchmark for the free-form generations of large vision-language models.
\newblock In \emph{2024 IEEE/CVF Conference on Computer Vision and Pattern Recognition (CVPR)}, pages 27218--27228. IEEE.

\bibitem[{Kogilathota et~al.(2026)Kogilathota, EG, Sun, and Zhou}]{kogilathota2026halp}
Sai~Akhil Kogilathota, Sripadha~Vallabha EG, Luzhe Sun, and Jiawei Zhou. 2026.
\newblock Halp: Detecting hallucinations in vision-language models without generating a single token.
\newblock In \emph{Proceedings of the 19th Conference of the European Chapter of the Association for Computational Linguistics (Volume 1: Long Papers)}, pages 6067--6085.

\bibitem[{Li et~al.(2023)Li, Du, Zhou, Wang, Zhao, and Wen}]{li2023pope}
Yifan Li, Yifan Du, Kun Zhou, Jinpeng Wang, Xin Zhao, and Ji-Rong Wen. 2023.
\newblock Evaluating object hallucination in large vision-language models.
\newblock In \emph{Proceedings of the 2023 conference on empirical methods in natural language processing}, pages 292--305.

\bibitem[{Lin et~al.(2014)Lin, Maire, Belongie, Hays, Perona, Ramanan, Doll{\'a}r, and Zitnick}]{lin2014coco}
Tsung-Yi Lin, Michael Maire, Serge Belongie, James Hays, Pietro Perona, Deva Ramanan, Piotr Doll{\'a}r, and C~Lawrence Zitnick. 2014.
\newblock Microsoft coco: Common objects in context.
\newblock In \emph{European conference on computer vision}, pages 740--755. Springer.

\bibitem[{Liu et~al.(2024{\natexlab{a}})Liu, Lin, Li, Wang, Yacoob, and Wang}]{liu2024lrv}
Fuxiao Liu, Kevin Lin, Linjie Li, Jianfeng Wang, Yaser Yacoob, and Lijuan Wang. 2024{\natexlab{a}}.
\newblock Mitigating hallucination in large multi-modal models via robust instruction tuning.
\newblock In \emph{International Conference on Learning Representations}, volume 2024, pages 57689--57733.

\bibitem[{Liu et~al.(2024{\natexlab{b}})Liu, Li, Li, and Lee}]{liu2024llava15}
Haotian Liu, Chunyuan Li, Yuheng Li, and Yong~Jae Lee. 2024{\natexlab{b}}.
\newblock Improved baselines with visual instruction tuning.
\newblock In \emph{2024 IEEE/CVF Conference on Computer Vision and Pattern Recognition (CVPR)}, pages 26286--26296. IEEE.

\bibitem[{Liu et~al.(2025)Liu, Fu, Xie, Xie, Sun, Lian, Kang, and Li}]{liu2025phd}
Jiazhen Liu, Yuhan Fu, Ruobing Xie, Runquan Xie, Xingwu Sun, Fengzong Lian, Zhanhui Kang, and Xirong Li. 2025.
\newblock Phd: A chatgpt-prompted visual hallucination evaluation dataset.
\newblock In \emph{2025 IEEE/CVF Conference on Computer Vision and Pattern Recognition (CVPR)}, pages 19857--19866. IEEE.

\bibitem[{Lovenia et~al.(2024)Lovenia, Dai, Cahyawijaya, Ji, and Fung}]{lovenia2023nope}
Holy Lovenia, Wenliang Dai, Samuel Cahyawijaya, Ziwei Ji, and Pascale Fung. 2024.
\newblock Negative object presence evaluation (nope) to measure object hallucination in vision-language models.
\newblock In \emph{Proceedings of the 3rd Workshop on Advances in Language and Vision Research (ALVR)}, pages 37--58.

\bibitem[{Min et~al.(2023)Min, Krishna, Lyu, Lewis, Yih, Koh, Iyyer, Zettlemoyer, and Hajishirzi}]{min2023factscore}
Sewon Min, Kalpesh Krishna, Xinxi Lyu, Mike Lewis, Wen-tau Yih, Pang Koh, Mohit Iyyer, Luke Zettlemoyer, and Hannaneh Hajishirzi. 2023.
\newblock Factscore: Fine-grained atomic evaluation of factual precision in long form text generation.
\newblock In \emph{Proceedings of the 2023 conference on empirical methods in natural language processing}, pages 12076--12100.

\bibitem[{Morante and Blanco(2012)}]{morante2012sem}
Roser Morante and Eduardo Blanco. 2012.
\newblock * sem 2012 shared task: Resolving the scope and focus of negation.
\newblock In \emph{* SEM 2012: The First Joint Conference on Lexical and Computational Semantics--Volume 1: Proceedings of the main conference and the shared task, and Volume 2: Proceedings of the Sixth International Workshop on Semantic Evaluation (SemEval 2012)}, pages 265--274.

\bibitem[{Morante and Daelemans(2009)}]{morante2009metalearning}
Roser Morante and Walter Daelemans. 2009.
\newblock A metalearning approach to processing the scope of negation.
\newblock In \emph{Proceedings of the thirteenth conference on computational natural language learning (CoNLL-2009)}, pages 21--29.

\bibitem[{Park and Li(2026)}]{park2026glsim}
Seongheon Park and Sharon Li. 2026.
\newblock Glsim: Detecting object hallucinations in lvlms via global-local similarity.
\newblock \emph{Advances in Neural Information Processing Systems}, 38:37747--37776.

\bibitem[{Prabhakaran et~al.(2025)Prabhakaran, Aggarwal, Verma, Swamy, and Saladi}]{prabhakaran2025vade}
Vishnu Prabhakaran, Purav Aggarwal, Vinay~Kumar Verma, Gokul Swamy, and Anoop Saladi. 2025.
\newblock Vade: Visual attention guided hallucination detection and elimination.
\newblock In \emph{Findings of the Association for Computational Linguistics: ACL 2025}, pages 14949--14965.

\bibitem[{Rohrbach et~al.(2018)Rohrbach, Hendricks, Burns, Darrell, and Saenko}]{rohrbach2018object}
Anna Rohrbach, Lisa~Anne Hendricks, Kaylee Burns, Trevor Darrell, and Kate Saenko. 2018.
\newblock Object hallucination in image captioning.
\newblock In \emph{Proceedings of the 2018 conference on empirical methods in natural language processing}, pages 4035--4045.

\bibitem[{Rudinger et~al.(2018)Rudinger, White, and Van~Durme}]{rudinger2018factuality}
Rachel Rudinger, Aaron~Steven White, and Benjamin Van~Durme. 2018.
\newblock Neural models of factuality.
\newblock In \emph{Proceedings of the 2018 Conference of the North American Chapter of the Association for Computational Linguistics: Human Language Technologies, Volume 1 (Long Papers)}, pages 731--744.

\bibitem[{Saur{\'\i} and Pustejovsky(2009)}]{sauri2009factbank}
Roser Saur{\'\i} and James Pustejovsky. 2009.
\newblock Factbank: a corpus annotated with event factuality.
\newblock \emph{Language resources and evaluation}, 43(3):227--268.

\bibitem[{Saur{\'\i} and Pustejovsky(2012)}]{sauri2012sure}
Roser Saur{\'\i} and James Pustejovsky. 2012.
\newblock Are you sure that this happened? assessing the factuality degree of events in text.
\newblock \emph{Computational linguistics}, 38(2):261--299.

\bibitem[{Sun et~al.(2024)Sun, Shen, Cao, Liu, Li, Shen, Gan, Gui, Wang, Yang et~al.}]{sun2023mmhal}
Zhiqing Sun, Sheng Shen, Shengcao Cao, Haotian Liu, Chunyuan Li, Yikang Shen, Chuang Gan, Liangyan Gui, Yu-Xiong Wang, Yiming Yang, et~al. 2024.
\newblock Aligning large multimodal models with factually augmented rlhf.
\newblock In \emph{Findings of the association for computational linguistics: ACL 2024}, pages 13088--13110.

\bibitem[{Vincze et~al.(2008)Vincze, Szarvas, Farkas, M{\'o}ra, and Csirik}]{vincze2008bioscope}
Veronika Vincze, Gy{\"o}rgy Szarvas, Rich{\'a}rd Farkas, Gy{\"o}rgy M{\'o}ra, and J{\'a}nos Csirik. 2008.
\newblock The bioscope corpus: biomedical texts annotated for uncertainty, negation and their scopes.
\newblock \emph{BMC bioinformatics}, 9(Suppl 11):S9.

\bibitem[{Wang et~al.(2023)Wang, Wang, Xu, Zhang, Gu, Jia, Wang, Xu, Yan, Zhang et~al.}]{wang2023amber}
Junyang Wang, Yuhang Wang, Guohai Xu, Jing Zhang, Yukai Gu, Haitao Jia, Jiaqi Wang, Haiyang Xu, Ming Yan, Ji~Zhang, et~al. 2023.
\newblock Amber: An llm-free multi-dimensional benchmark for mllms hallucination evaluation.
\newblock \emph{arXiv preprint arXiv:2311.07397}.

\bibitem[{Wu et~al.(2024)Wu, Ji, Huang, Li, Wu, Sun, and Ji}]{wu2024rbench}
Mingrui Wu, Jiayi Ji, Oucheng Huang, Jiale Li, Yuhang Wu, Xiaoshuai Sun, and Rongrong Ji. 2024.
\newblock Evaluating and analyzing relationship hallucinations in large vision-language models.
\newblock \emph{arXiv preprint arXiv:2406.16449}.

\bibitem[{Zhang et~al.(2025)Zhang, Xie, Sun, Huang, Chen, Kang, Wang, and Wang}]{zhang2025dhcp}
Yudong Zhang, Ruobing Xie, Xingwu Sun, Yiqing Huang, Jiansheng Chen, Zhanhui Kang, Di~Wang, and Yu~Wang. 2025.
\newblock Dhcp: Detecting hallucinations by cross-modal attention pattern in large vision-language models.
\newblock In \emph{Proceedings of the 33rd ACM International Conference on Multimedia}, pages 3555--3564.

\bibitem[{Zhou et~al.(2024)Zhou, Cui, Yoon, Zhang, Deng, Finn, Bansal, and Yao}]{zhou2024analyzing}
Yiyang Zhou, Chenhang Cui, Jaehong Yoon, Linjun Zhang, Zhun Deng, Chelsea Finn, Mohit Bansal, and Huaxiu Yao. 2024.
\newblock Analyzing and mitigating object hallucination in large vision-language models.
\newblock In \emph{International Conference on Learning Representations}, volume 2024, pages 56969--56998.

\bibitem[{Zhu et~al.(2025)Zhu, Wang, Chen, Liu, Ye, Gu, Tian, Duan, Su, Shao et~al.}]{zhu2025internvl3}
Jinguo Zhu, Weiyun Wang, Zhe Chen, Zhaoyang Liu, Shenglong Ye, Lixin Gu, Hao Tian, Yuchen Duan, Weijie Su, Jie Shao, et~al. 2025.
\newblock Internvl3: Exploring advanced training and test-time recipes for open-source multimodal models.
\newblock \emph{arXiv preprint arXiv:2504.10479}.

\bibitem[{Zimmermann(1993)}]{zimmermann1993opacity}
Thomas~Ede Zimmermann. 1993.
\newblock On the proper treatment of opacity in certain verbs.
\newblock \emph{Natural language semantics}, 1(2):149--179.

\bibitem[{Zohrabi et~al.(2026)Zohrabi, Hasani, Gupta, Baghshah, Rohrbach, and Rohrbach}]{zohrabi2026haloprobe}
Reihaneh Zohrabi, Hosein Hasani, Akshita Gupta, Mahdieh~Soleymani Baghshah, Anna Rohrbach, and Marcus Rohrbach. 2026.
\newblock Haloprobe: Bayesian detection and mitigation of object hallucinations in vision-language models.
\newblock \emph{arXiv preprint arXiv:2604.06165}.

\end{thebibliography}

\clearpage

\appendix

\section{Category Vocabulary}
\label{app:vocab}
Candidate mentions are found by a case-insensitive, whole-word regular expression over the forms below. For each
COCO category we list the additional surface forms beyond its name. Regular plurals of single-word forms are added
automatically (\emph{-s}, \emph{-es}, \emph{-ies}); irregular plurals are listed explicitly. The category
\emph{orange} is matched but excluded from \bench{}.

{\footnotesize\raggedright
\textbf{person}: people, man, woman, men, women, boy, girl, child, children, kid, guy, lady, human, someone, pedestrian, passerby; \textbf{bicycle}: bike, cyclist; \textbf{car}: sedan, vehicle, automobile, taxi, cab, suv, auto; \textbf{motorcycle}: motorbike, moped, scooter; \textbf{airplane}: aeroplane, plane, jet, aircraft; \textbf{bus}; \textbf{train}: locomotive; \textbf{truck}: lorry, pickup; \textbf{boat}: ship, vessel, canoe, kayak; \textbf{traffic light}: stoplight, signal light; \textbf{fire hydrant}: hydrant; \textbf{stop sign}; \textbf{parking meter}; \textbf{bench}; \textbf{bird}: sparrow, pigeon, crow, eagle, parrot, duck, goose, swan, seagull; \textbf{cat}: kitten, kitty, feline; \textbf{dog}: puppy, canine; \textbf{horse}: pony, stallion, mare; \textbf{sheep}: lamb, ram, ewe; \textbf{cow}: cattle, calf, bull, ox; \textbf{elephant}; \textbf{bear}: grizzly; \textbf{zebra}; \textbf{giraffe}; \textbf{backpack}: rucksack, knapsack; \textbf{umbrella}: parasol; \textbf{handbag}: purse, tote, satchel; \textbf{tie}: necktie, bowtie; \textbf{suitcase}: luggage, baggage; \textbf{frisbee}: disc; \textbf{skis}: ski; \textbf{snowboard}; \textbf{sports ball}: soccer ball, tennis ball, basketball, baseball, football, volleyball; \textbf{kite}; \textbf{baseball bat}; \textbf{baseball glove}: baseball mitt; \textbf{skateboard}; \textbf{surfboard}; \textbf{tennis racket}: tennis racquet, racket, racquet; \textbf{bottle}; \textbf{wine glass}: wineglass, goblet; \textbf{cup}: mug; \textbf{fork}; \textbf{knife}: knives; \textbf{spoon}; \textbf{bowl}; \textbf{banana}; \textbf{apple}; \textbf{sandwich}: sub; \textbf{orange}; \textbf{broccoli}; \textbf{carrot}; \textbf{hot dog}: hotdog; \textbf{pizza}; \textbf{donut}: doughnut; \textbf{cake}: cupcake; \textbf{chair}: armchair, stool; \textbf{couch}: sofa, loveseat; \textbf{potted plant}: houseplant, plant pot; \textbf{bed}; \textbf{dining table}: dinner table, kitchen table; \textbf{toilet}: commode; \textbf{tv}: television, tv set; \textbf{laptop}: notebook computer; \textbf{mouse}: computer mouse; \textbf{remote}: remote control; \textbf{keyboard}; \textbf{cell phone}: cellphone, mobile phone, smartphone, phone; \textbf{microwave}; \textbf{oven}: stove; \textbf{toaster}; \textbf{sink}: washbasin; \textbf{refrigerator}: fridge; \textbf{book}; \textbf{clock}; \textbf{vase}; \textbf{scissors}; \textbf{teddy bear}: teddy; \textbf{hair drier}: hair dryer, hairdryer, blow dryer; \textbf{toothbrush}.\par}

\section{Detectors}
\label{app:detectors}

\paragraph{Common setup.} Table \ref{tab:detectors} summarises each detector’s signal and classifier. Every detector scores a single object mention. All internal signals are read from one
teacher-forced forward pass of the captioning model over the exact prompt and caption it generated (Qwen-style chat
template for the newer models). The \emph{object token} is the first sub-token of the object word at the labelled
occurrence; because a caption token is generated by the query one position earlier, generation-time signals are read
at that preceding position, as they would be during decoding. Signals that the original methods average over a span
are averaged over a fixed window of the four caption tokens ending at the object token (W4), the same for every class,
so that no feature depends on the trigger phrase returned by the labelling judge. Higher scores mean ``more likely
hallucinated''; for scores defined the other way round we negate them.

\paragraph{Training-free detectors.} These compute a score directly from the model, with no parameters fitted to our
labels. We use each method's published definition and default settings (layers, top-$k$, weights) and adapt only the
layer indices to model depth by a fixed rule; nothing is selected on our data. The only labelled data they see is the
calibration set, used to set the flagging threshold, exactly as for trained detectors.
They therefore cannot change with \fm{} training data, and appear as a single value in Table~\ref{tab:main}.

\paragraph{Trained detectors.} These fit a classifier on labelled mentions of the fit set (label 1: hallucinated;
label 0: concrete, plus any mixed-in \fm{} mentions). We use each method's published classifier: a logistic regression
for Lookback Lens and for our hidden-state probes, and the papers' MLPs for DHCP and HALP.

\begin{table*}[t]
\centering\small\setlength{\tabcolsep}{4pt}
\resizebox{\textwidth}{!}{%
\begin{tabular}{p{2.6cm} p{2.4cm} p{6.4cm} p{4.6cm}}
\toprule
Detector & Source & Signal & Score / classifier \\
\midrule
\multicolumn{4}{l}{\textit{Training-free}} \\
SVAR & \citet{jiang2025devils} & total attention to image tokens at the object's generation step, head-mean, summed over the
  middle layers (5--18 of 32, scaled to depth) & $-$attention (low image attention = hallucination) \\
VADE (VisAttnAgg) & \citet{prabhakaran2025vade} & mean attention to image tokens over layers, heads and the W4 window & $-$attention \\
Internal Confidence & \citet{jiang2025interpreting} & logit-lens probability of the object word at image tokens, max over layers and image tokens (top-50 softmax) & $-$confidence \\
GLSim & \citet{park2026glsim} & $0.6\times$ global + $0.4\times$ local similarity between the object's representation and the image (official defaults: text layer $L{-}1$, image layer $L$, $k{=}32$) & $-$similarity \\
PAS & \citet{hoang2026pas} & attention of the object token to previously generated tokens, layer 0 (paper default) & score as defined \\
NLL / entropy & \citet{zhou2024analyzing} & negative log-probability of the object token / entropy of the next-token distribution & score as defined \\
\midrule
\multicolumn{4}{l}{\textit{Trained}} \\
Lookback Lens & \citet{chuang2024lookback} & lookback ratio (attention to the prompt vs.\ to generated tokens) of every head, W4 & logistic regression (official) \\
Hidden state (mid / last) & ours & hidden state at layer $L/2$ / the last layer, mean over W4 & standardised, class-balanced logistic regression \\
Text-only LM & ours & as hidden state (mid), but the caption is read by the model's language model without the image & as hidden state \\
DHCP & \citet{zhang2025dhcp} & attention to every image token at the object token, all layers and heads, image grid pooled to $6{\times}6$ & MLP 512--128, 100 epochs, batch 1024, class-balanced sampling \\
HALP & \citet{kogilathota2026halp} & hidden state at the last prompt token (before any caption), last layer; one per image & MLP 512--256--128 (BN, dropout 0.3), 50 epochs, batch 32 \\
\bottomrule
\end{tabular}}
\caption{Detectors. $L$: number of layers of the captioning model; W4: the four caption tokens ending at the object
token. All MLPs use Adam (learning rate $10^{-3}$) and binary cross-entropy; DHCP inputs are standardised with fit-set
statistics.}
\label{tab:detectors}
\end{table*}

\paragraph{Adaptations to the newer models.} Layer-indexed settings are scaled to the model's depth: the middle layer
is $L/2$ (14 of 28 for Qwen2.5-VL-7B, 18 of 36 for the 36-layer models), SVAR sums layers 5--18 of 32 scaled
proportionally, GLSim uses layers $L{-}1$ and $L$, and HALP reads the last layer. For DHCP, the variable number of
image tokens of Qwen-style encoders is pooled into the same $6{\times}6$ grid. Lookback Lens takes as context
everything before the assistant turn, as in its official code.

\paragraph{Training and evaluation protocol.}
\label{app:protocol}
For each model and each of three seeds, the images are split into a fit (53.3\%), a calibration (13.3\%) and a
held-out (33.3\%) set, with all mentions of an image in the same set. Trained detectors are fit on the fit set with
the labels above; \fm{} mentions are added as label-0 examples at the mixing levels of Table~\ref{tab:mixing-side},
drawn in a fixed random order. For every detector, trained or not, the threshold is the 90th percentile of its scores
on the calibration set's concrete mentions, so that it flags 10\% of them. On the held-out set we report the share of
\fm{} mentions flagged (FPR), the share of hallucinations flagged (TPR) and AUCs, averaged over the seeds. All experiments were run on a single NVIDIA H100 GPU.

\section{Labelling Prompts}
\label{app:prompt}

Every mention is labelled in three LLM steps, all at temperature 0 with JSON output, plus deterministic rules.
Candidate mentions are found with a regular expression over the 80 COCO class names, their synonyms and plurals.
The LLM judges see only the caption text, never the image or its annotations. The prompts are reproduced verbatim
and use our earlier internal terminology: \textsc{functional} is the label for a \fm{} mention, and step~2 asks for
a five-way subtype from which the four \fm{} categories are derived (Table~\ref{tab:category-map}).

\paragraph{Step 1: usage (GPT-4o; Figure~\ref{fig:prompt-usage}).} For each candidate mention, the judge decides
whether the caption refers to the concept at all and, if so, whether it asserts that the object is present. Six
captions are labelled per call.

\paragraph{Step 2: category (GPT-4o; Figure~\ref{fig:prompt-category}, top).} Each \fm{} mention (judged
\textsc{functional}) is assigned the mechanism by which it avoids asserting the object, given the full caption and
the trigger phrase from step~1.

\paragraph{Step 3: compounds (GPT-4o-mini; Figure~\ref{fig:prompt-category}, bottom).} Each \fm{}
mention is checked for being part of a larger expression with a different meaning (\emph{hot dog}, \emph{teddy bear}).

\paragraph{Deterministic rules.} A mention judged \textsc{existential} is \emph{concrete} if its class appears in the
image's COCO instance annotations and \emph{hallucinated} otherwise; mentions judged \textsc{ambiguous} or not
actually referring to the concept are discarded. Mentions judged \textsc{functional} are \fm{}. The step-2 prompt
uses the original five-way scheme; the four categories of Section~\ref{sec:def} are derived from its output by the
rules in Table~\ref{tab:category-map}, which merge purpose and intensional into one category, add negation, and
drop the remaining generic mentions.
Negation is detected by a cue word in the step-1 trigger phrase (\emph{no, not, none, without, absent, absence,
lack(s/ing), nor, neither, devoid, n't}) and overrides the step-2 label except for depictive mentions (a sign that
says ``no dogs'' depicts content). Negations are therefore not discarded: the step-2 prompt lists ``no people are
visible'' as an example of \emph{generic}, but every generic mention with a negation cue is reassigned to negation by
this rule, and only generic mentions without one (kind-level or habitual references such as ``a road where cars
usually drive'') are dropped. The concept \emph{orange} is excluded throughout, because the step-1 judge often
accepted the colour as a mention of the fruit.

\begin{table}[t]
\centering\small\setlength{\tabcolsep}{4pt}
\resizebox{\columnwidth}{!}{%
\begin{tabular}{ll}
\toprule
Step-2 / step-3 output & Category \\
\midrule
Any subtype except depictive, with a negation cue & Negation \\
Hedging & Hedged \\
Depictive (with or without a negation cue) & Depictive \\
Purpose, intensional & Purpose \\
Generic without a negation cue, other & excluded \\
Compound (step 3) & excluded \\
\bottomrule
\end{tabular}}
\caption{Mapping from the step-2 subtype and the step-3 compound check to the four \fm{} categories.}
\label{tab:category-map}
\end{table}

\begin{figure*}[p]
\centering
\textbf{System:}
\begin{lstlisting}[style=prompt]
You are a precise linguistic analyst. Output ONLY valid JSON -- no markdown, no text outside the JSON.
\end{lstlisting}
\textbf{User:}
\begin{lstlisting}[style=prompt]
You will analyze {N} SEPARATE captions below. Each has its own caption index and its own list of target concepts -- treat each one completely independently; do not let one caption's content influence another's classification.

For EACH target concept in EACH caption, analyze how that concept is used in ITS OWN caption.

For each concept:
STEP 1 -- Mention check: does the caption actually refer to this concept (including synonyms)?  If not, concept_mentioned = false.

STEP 2 -- If mentioned, classify the mention TYPE (based on wording only, NOT on whether the object is really in any image):

  EXISTENTIAL -- the caption asserts this concept EXISTS as a physical object/entity in the depicted scene.
     e.g. "a person is standing", "there are two cars parked", "a cat sits on the mat".

  FUNCTIONAL  -- the concept appears ONLY in a functional, purposive, relational, or hypothetical role, WITHOUT asserting it is physically present in this scene.
     e.g. "a bridge built for PEOPLE to cross"  (people = purpose)
          "a stool used for SITTING"            (no person asserted)
          "a road where CARS usually drive"     (no specific car asserted)
          "a rack designed to hold BOTTLES"     (no bottle asserted)
          "large enough to fit a TRUCK"         (hypothetical capacity)
          "no PEOPLE are visible"               (asserted absence)

  AMBIGUOUS   -- genuinely unclear, or used both ways.

Captions to analyze:

--- Caption index 0 ---
Caption: "{caption}"
Target concepts for this caption:
    - "{concept}"
...

Return ONLY this JSON object, with one entry in "captions" per caption index above (do not skip any index, even if a concept is not mentioned):
{
  "captions": [
    {
      "index": <caption index integer>,
      "results": [
        {
          "concept": "<one of that caption's target concepts>",
          "concept_mentioned": true | false,
          "trigger_phrase": "exact substring from caption, or null",
          "mention_type": "EXISTENTIAL" | "FUNCTIONAL" | "AMBIGUOUS" | null,
          "reasoning": "3-6 words max, e.g. 'purpose clause, no object asserted'",
          "confidence": "HIGH" | "MEDIUM" | "LOW"
        }
      ]
    }
  ]
}
\end{lstlisting}
\caption{Step-1 prompt (GPT-4o, temperature 0): mention check and usage. \{N\}, \{caption\} and \{concept\} are
filled in per call; six captions per call.}
\label{fig:prompt-usage}
\end{figure*}

\begin{figure*}[p]
\centering
\textbf{System (steps 2 and 3):}
\begin{lstlisting}[style=prompt]
You are a precise linguistic annotator. Output ONLY valid JSON.
\end{lstlisting}
\textbf{Step 2, user (GPT-4o):}
\begin{lstlisting}[style=prompt]
Classify the linguistic mechanism by which the following object mention avoids asserting the object's existence in an image.

Full caption: "{caption}"
Trigger span: "{trigger_phrase}"
Concept word: {concept}

Categories:
- purpose: The object is mentioned as the intended user/beneficiary of a purpose clause (e.g. 'a cup for holding toothbrushes', 'a bench for people to sit on'). The construction explains what an anchor object is FOR.
- intensional: The object appears inside a non-veridical/intensional context governed by a verb like want, wait for, hope, look for, expect (e.g. 'waiting for the train', 'looking for a taxi'). The mention does not entail the object's presence.
- hedging: The mention is marked with epistemic uncertainty (possibly, likely, maybe, seems, appears) about whether the object is present (e.g. 'possibly a bird', 'what appears to be a dog').
- depictive: The object is mentioned as depicted content (a sign, picture, logo, text, or symbol) rather than a real physical instance (e.g. 'a sign that says no dogs', 'a picture of a cat').
- generic: The object is referred to as a kind or in a generic/habitual statement rather than a specific instance in the scene (e.g. 'a road where cars usually drive', 'a place popular with tourists', 'no people are visible').

If the span could fit multiple categories, pick the SINGLE best/primary one. If none fit well, use "other".

Output JSON: {"subtype": "purpose|intensional|hedging|depictive|generic|other", "confidence": "high|medium|low", "reason": "one short phrase"}
\end{lstlisting}
\textbf{Step 3, user (GPT-4o-mini):}
\begin{lstlisting}[style=prompt]
In the phrase below, a word or phrase matching the concept "{concept}" was matched. Determine whether this match is actually part of a LARGER compound expression whose overall meaning is different from "{concept}" alone (e.g., "hot dog" means a food item, not "dog"; "toilet bowl" means a fixture, not a "bowl"; "orange juice" is a drink, not the fruit "orange").

Phrase: "{trigger_phrase}"
Matched concept: {concept}

If the match is part of such a compound (the combined phrase has a distinct, different meaning from the bare concept), identify the FULL compound expression.
If the match is NOT part of a compound (concept is used in its ordinary/standalone sense, possibly with an ordinary adjective like "large orange cat" where orange is a colour, or "red car"), say so.

Output JSON: {"is_compound": true/false, "compound_phrase": "the full compound or null", "compound_meaning": "brief gloss of what the compound refers to, or null"}
\end{lstlisting}
\caption{Prompts for the category of a \fm{} mention (step 2) and the compound check (step 3), temperature 0.
\{caption\}, \{trigger\_phrase\} and \{concept\} are filled in per mention.}
\label{fig:prompt-category}
\end{figure*}

\section{Teacher-Forced Control}
\label{app:teacher-forced}
Table~\ref{tab:mixing-optA} repeats the mixing experiment of Table~\ref{tab:mixing-side} with every model reading
LLaVA-1.5's captions instead of its own (\S\ref{sec:res-mixing}).

\begin{table*}[t]
\centering\small\setlength{\tabcolsep}{2.5pt}
\resizebox{\textwidth}{!}{%
\begin{tabular}{l|ccccc|ccccc|ccccc|ccccc|ccccc}
\toprule
 & \multicolumn{5}{c|}{LLaVA-1.5-7B} & \multicolumn{5}{c|}{Qwen2.5-VL-7B} & \multicolumn{5}{c|}{Qwen2.5-VL-3B} & \multicolumn{5}{c|}{InternVL3-8B} & \multicolumn{5}{c}{LLaVA-OV2-8B} \\
Detector & 0 & nat. & 10 & 30 & max & 0 & nat. & 10 & 30 & max & 0 & nat. & 10 & 30 & max & 0 & nat. & 10 & 30 & max & 0 & nat. & 10 & 30 & max \\
\midrule
\multicolumn{26}{l}{\textit{\Fm{} FPR (\%) $\downarrow$}} \\
Lookback Lens       & 40 & 17 & 4 & 2 & 1 & 44 & 9 & 3 & 2 & 1 & 34 & 10 & 2 & 1 & 1 & 31 & 10 & 3 & 2 & 1 & 38 & 10 & 3 & 2 & 1 \\
Hidden state (mid)  & 52 & 23 & 6 & 3 & 2 & 47 & 23 & 7 & 4 & 2 & 50 & 23 & 9 & 4 & 3 & 55 & 29 & 11 & 4 & 3 & 44 & 16 & 6 & 4 & 3 \\
Hidden state (last) & 34 & 17 & 5 & 3 & 2 & 34 & 19 & 7 & 4 & 2 & 32 & 20 & 6 & 3 & 2 & 39 & 21 & 9 & 5 & 3 & 39 & 20 & 8 & 4 & 2 \\
DHCP                & 18 & 11 & 5 & 2 & 1 & 39 & 13 & 3 & 2 & 1 & 38 & 10 & 3 & 2 & 1 & 29 & 10 & 4 & 2 & 2 & 27 & 12 & 3 & 1 & 1 \\
Text-only LM        & 46 & 18 & 5 & 2 & 1 & 36 & 14 & 6 & 3 & 1 & 42 & 21 & 7 & 3 & 2 & 38 & 18 & 6 & 3 & 2 & 41 & 15 & 6 & 3 & 2 \\
\cmidrule(lr){1-26}
HALP                & 22 & 15 & 12 & 8 & 7 & 17 & 12 & 15 & 10 & 7 & 15 & 17 & 15 & 13 & 8 & 16 & 15 & 15 & 8 & 6 & 17 & 15 & 9 & 9 & 5 \\
\midrule
\multicolumn{26}{l}{\textit{Hallucination TPR (\%) $\uparrow$}} \\
Lookback Lens       & 54 & 54 & 52 & 50 & 49 & 51 & 52 & 52 & 52 & 52 & 46 & 46 & 47 & 46 & 45 & 48 & 47 & 44 & 43 & 42 & 51 & 51 & 50 & 50 & 47 \\
Hidden state (mid)  & 49 & 50 & 50 & 51 & 52 & 61 & 62 & 60 & 61 & 61 & 59 & 60 & 62 & 61 & 62 & 65 & 62 & 62 & 62 & 63 & 62 & 63 & 63 & 66 & 63 \\
Hidden state (last) & 46 & 43 & 42 & 43 & 43 & 56 & 56 & 54 & 54 & 56 & 50 & 50 & 50 & 46 & 47 & 61 & 59 & 60 & 58 & 59 & 52 & 53 & 52 & 51 & 50 \\
DHCP                & 43 & 37 & 40 & 42 & 43 & 48 & 47 & 43 & 45 & 46 & 43 & 41 & 44 & 44 & 43 & 43 & 45 & 47 & 43 & 42 & 44 & 46 & 42 & 46 & 39 \\
Text-only LM        & 35 & 35 & 36 & 35 & 36 & 33 & 31 & 32 & 31 & 31 & 30 & 32 & 32 & 33 & 33 & 34 & 33 & 34 & 33 & 34 & 32 & 34 & 36 & 34 & 35 \\
\cmidrule(lr){1-26}
HALP                & 14 & 12 & 12 & 10 & 12 & 15 & 13 & 16 & 17 & 16 & 13 & 15 & 15 & 16 & 12 & 12 & 12 & 16 & 12 & 11 & 13 & 16 & 12 & 13 & 9 \\
\bottomrule
\end{tabular}}
\caption{\Fm{} FPR (top) and hallucination TPR (bottom) of the trained detectors when every model reads LLaVA-1.5's
captions instead of its own (the LLaVA-1.5 columns repeat Table~\ref{tab:mixing-side}). Columns as in
Table~\ref{tab:mixing-side}; \emph{nat.} is LLaVA-1.5's natural share (about 20 examples) and \emph{max} is 77\% for
every model, since all read the same captions.}
\label{tab:mixing-optA}
\end{table*}

\section{Text Filter}
\label{app:filter}
\paragraph{Setup.}
The filter is a separate class-balanced logistic regression
trained to distinguish \fm{} mentions (label 1) from
concrete and hallucinated mentions (label 0). It uses
the same features as the text-only probe: middle-layer
hidden states from the language model without the image,
averaged over the W4 window. Training uses the fit split's
concrete and hallucinated mentions and the natural-share
budget of \fm{} mentions (20 for LLaVA-1.5, 76--108 for
the newer models). The decision threshold is fixed at
0.5, without tuning. Training, calibration and test
splits are disjoint by image. After filtering, each
detector's threshold is recalibrated on the remaining
concrete calibration mentions to maintain 10\% concrete
FPR. Hallucinations removed (\%) denotes the fraction
of held-out hallucinated mentions incorrectly filtered.

Table~\ref{tab:filter} gives the per-category recall of the text filter and its effect on each detector group
(\S\ref{sec:res-filter}).

\begin{table*}[t]
\centering\small\setlength{\tabcolsep}{4pt}
\resizebox{\textwidth}{!}{%
\begin{tabular}{l ccccc c cc cc}
\toprule
 & \multicolumn{5}{c}{Filter recall on \fm{} (\%)} & Halluc. & \multicolumn{2}{c}{Training-free FPR} & \multicolumn{2}{c}{Trained probes FPR} \\
\cmidrule(lr){2-6}\cmidrule(lr){8-9}\cmidrule(lr){10-11}
Model & Hedg. & Dep. & Purp. & Neg. & All & removed (\%) & no filter & filter & mixing & + filter \\
\midrule
LLaVA-1.5-7B         & 67 & 17 & 79 & 80 & 64 & 0.2 & 14--57 & 2--11  & 11--23 & 4--13  \\
Qwen2.5-VL-7B        & 53 & 70 & 65 & 88 & 62 & 2.5 & 21--53 & 7--19  & 17--34 & 14--18 \\
Qwen2.5-VL-3B        & 52 & 67 & 65 & 99 & 64 & 3.1 & 21--53 & 6--16  & 22--31 & 14--17 \\
InternVL3-8B         & 58 & 72 & 8  & 99 & 69 & 2.1 & 16--46 & 4--13  & 16--37 & 11--16 \\
LLaVA-OneVision-2-8B & 50 & 61 & 76 & 98 & 62 & 3.8 & 16--57 & 4--18  & 26--46 & 19--22 \\
\bottomrule
\end{tabular}}
\caption{Text filter placed in front of each detector. Left: share of \fm{} mentions it catches and share of
hallucinations it wrongly removes. Right: range of \fm{} FPR (\%) over the seven training-free detectors and over
the five trained probes mixed at the natural share, without and with the filter.}
\label{tab:filter}
\end{table*}

\section{THRONE: Voting Rule and Judge}
\label{app:throne}

\paragraph{Setup.} THRONE~\cite{kaul2024throne} asks a language model, for each object class, whether a caption
implies that the object is in the image, and counts a class as hallucinated if the answer is yes and the object is
not annotated. We run its released prompts on the mentions of each model's evaluation set
(Table~\ref{tab:dataset_counts}): the first three question formats, its instruction, captions truncated to 1{,}024
characters and greedy decoding, asking only about the class that was mentioned. THRONE's judges are three Flan-T5
models (base, large and XL), giving nine yes/no votes per mention. A mention counts as hallucinated if the object is
judged present and is not in the COCO annotations.

\paragraph{Voting rule.} THRONE's default rule is unanimous: an object is present if all nine votes are yes, absent if
none are, and \emph{disqualified} otherwise, and disqualified mentions are dropped from its metric. On our mentions this
rule disqualifies 61\% of the newer models' \fm{} mentions (68\% of hedged ones), against 6\% of concrete mentions
(Table~\ref{tab:throne-judges}). It therefore avoids scoring most \fm{} mentions rather than scoring them correctly: on
the mentions it does score, it counts 46\% of \fm{} mentions as hallucinated, as many as the majority vote (51\%). It
also disqualifies 22\% of the newer models' real hallucinations. We therefore report a majority vote, which gives every
mention a verdict.

\paragraph{A stronger judge.} To check that THRONE's failure on \fm{} mentions is not specific to its small Flan-T5
judges, we replace them with Llama-3.1-8B-Instruct, keeping the same prompts (sent as the user turn of its chat
template) and taking a vote as yes if the answer begins with ``yes''. Three questions give three votes; the majority
vote needs two. The stronger judge reduces but does not remove the problem: it still counts 41\% of the newer models'
\fm{} mentions as hallucinated, including half of their hedged mentions (50\%) and almost half of their depictive ones
(47\%), while negation stays near zero for both judges. Its unanimous vote disqualifies fewer mentions than Flan-T5's,
but still mostly hedged ones (56\%), and on the mentions it scores it flags as many \fm{} mentions as its majority vote
(39\% vs.\ 41\%).

\begin{table*}[t]
\centering\small
\begin{tabular}{ll c cc ccccc c}
\toprule
 & & & \multicolumn{2}{c}{Disqualified} & \multicolumn{5}{c}{\Fm{} counted as halluc.} & L1.5 \\
\cmidrule(lr){4-5}\cmidrule(lr){6-10}
Judge & Vote & Recall & \Fm{} & Conc. & All & Hedg. & Dep. & Purp. & Neg. & All \\
\midrule
Flan-T5 & majority  & 98 & 0  & 0 & 51 & 57 & 60 & 38 & 3 & 55 \\
        & unanimous & 78 & 61 & 6 & 46 & 55 & 59 & 39 & 0 & 60 \\
\midrule
Llama   & majority  & 93 & 0  & 0 & 41 & 50 & 47 & 24 & 2 & 30 \\
        & unanimous & 86 & 41 & 2 & 39 & 50 & 48 & 28 & 2 & 28 \\
\bottomrule
\end{tabular}
\caption{THRONE with its Flan-T5 judges (nine votes) and with Llama-3.1-8B-Instruct (three votes), under majority and
unanimous voting, on the evaluation-set mentions of the four newer models (pooled; \emph{L1.5}: LLaVA-1.5). Recall:
hallucinations counted as hallucinated, disqualified ones counted as missed. Disqualified: share of \fm{} and concrete
mentions with split votes. \Fm{} columns: share counted as hallucinated among the mentions that receive a verdict. All
values in \%.}
\label{tab:throne-judges}
\end{table*}

\section{Position-Matched Evaluation}
\label{app:position}
Table~\ref{tab:position} compares each detector's concrete-vs-hallucination AUC before and after matching token
position (\S\ref{sec:res-eval}).

\begin{table*}[t]
\centering\small\setlength{\tabcolsep}{4pt}
\begin{tabular}{l cc cc cc cc cc}
\toprule
 & \multicolumn{2}{c}{LLaVA-1.5-7B} & \multicolumn{2}{c}{Qwen2.5-VL-7B} & \multicolumn{2}{c}{Qwen2.5-VL-3B} & \multicolumn{2}{c}{InternVL3-8B} & \multicolumn{2}{c}{LLaVA-OV2-8B} \\
\cmidrule(lr){2-3}\cmidrule(lr){4-5}\cmidrule(lr){6-7}\cmidrule(lr){8-9}\cmidrule(lr){10-11}
Detector & All & Pos. & All & Pos. & All & Pos. & All & Pos. & All & Pos. \\
\midrule
Position only       & 0.81 & 0.51 & 0.67 & 0.50 & 0.72 & 0.50 & 0.69 & 0.50 & 0.68 & 0.50 \\
PAS                 & 0.82 & 0.53 & 0.67 & 0.51 & 0.72 & 0.51 & 0.68 & 0.51 & 0.68 & 0.51 \\
SVAR                & 0.77 & 0.51 & 0.70 & 0.62 & 0.73 & 0.61 & 0.67 & 0.55 & 0.75 & 0.68 \\
VADE                & 0.76 & 0.48 & 0.59 & 0.57 & 0.67 & 0.60 & 0.54 & 0.54 & 0.60 & 0.59 \\
Internal Conf.      & 0.74 & 0.69 & 0.55 & 0.54 & 0.63 & 0.60 & 0.58 & 0.57 & 0.62 & 0.63 \\
GLSim               & 0.82 & 0.63 & 0.52 & 0.51 & 0.54 & 0.48 & 0.61 & 0.56 & 0.59 & 0.57 \\
NLL                 & 0.72 & 0.65 & 0.65 & 0.62 & 0.67 & 0.63 & 0.63 & 0.61 & 0.59 & 0.60 \\
Entropy             & 0.73 & 0.66 & 0.66 & 0.63 & 0.69 & 0.65 & 0.65 & 0.63 & 0.61 & 0.60 \\
\midrule
Lookback Lens       & 0.88 & 0.73 & 0.87 & 0.84 & 0.88 & 0.83 & 0.85 & 0.79 & 0.91 & 0.87 \\
Hidden state (mid)  & 0.87 & 0.76 & 0.90 & 0.87 & 0.89 & 0.85 & 0.88 & 0.83 & 0.92 & 0.89 \\
Hidden state (last) & 0.86 & 0.72 & 0.89 & 0.87 & 0.89 & 0.87 & 0.88 & 0.84 & 0.92 & 0.90 \\
DHCP                & 0.84 & 0.65 & 0.87 & 0.84 & 0.86 & 0.82 & 0.83 & 0.77 & 0.86 & 0.82 \\
Text-only LM        & 0.81 & 0.65 & 0.85 & 0.82 & 0.83 & 0.79 & 0.82 & 0.77 & 0.88 & 0.86 \\
\bottomrule
\end{tabular}
\caption{Concrete-vs-hallucination AUC on all held-out mentions (All) and on a subset matched for token position
(Pos.), for detectors built without \fm{} training data. Top: training-free detectors and position alone; bottom:
trained probes. Detectors whose AUC falls to about 0.5 once position is matched mostly measure position.}
\label{tab:position}
\end{table*}

\section{Mixing by Category}
\label{app:cat-mixing}
Table~\ref{tab:cat4-mixing-sub} breaks the mixing results of Table~\ref{tab:mixing-side} down by category for the
two strongest probes (\S\ref{sec:res-findings}).

\begin{table*}[t]
\centering\small
\begin{tabular}{llrccccc}
\toprule
Model & Category & $n$ & 0\% & nat. & 10\% & 30\% & max \\
\midrule
\multirow{4}{*}{LLaVA-1.5-7B} & Hedged & 197 & 64 / 64 & 30 / 25 & 5 / 6 & 3 / 3 & 3 / 2 \\
 & Depictive & 99 & 27 / 43 & 22 / 37 & 8 / 14 & 3 / 6 & 2 / 2 \\
 & Purpose & 211 & 24 / 46 & 5 / 18 & 2 / 3 & 1 / 2 & 1 / 1 \\
 & Negation & 46 & 45 / 48 & 4 / 8 & 1 / 1 & 1 / 0 & 0 / 0 \\
\midrule
\multirow{4}{*}{Qwen2.5-VL-7B} & Hedged & 312 & 71 / 63 & 39 / 29 & 39 / 29 & 21 / 17 & 10 / 10 \\
 & Depictive & 212 & 75 / 66 & 30 / 19 & 30 / 20 & 11 / 10 & 4 / 5 \\
 & Purpose & 55 & 39 / 38 & 12 / 17 & 11 / 17 & 3 / 9 & 2 / 6 \\
 & Negation & 34 & 29 / 28 & 7 / 9 & 7 / 9 & 5 / 5 & 3 / 2 \\
\midrule
\multirow{4}{*}{Qwen2.5-VL-3B} & Hedged & 267 & 81 / 64 & 46 / 34 & 46 / 33 & 25 / 20 & 9 / 10 \\
 & Depictive & 164 & 57 / 51 & 25 / 17 & 25 / 16 & 7 / 8 & 2 / 4 \\
 & Purpose & 70 & 59 / 42 & 16 / 17 & 16 / 17 & 9 / 12 & 4 / 9 \\
 & Negation & 72 & 37 / 14 & 1 / 1 & 1 / 1 & 0 / 1 & 0 / 0 \\
\midrule
\multirow{4}{*}{InternVL3-8B} & Hedged & 68 & 80 / 70 & 49 / 27 & 49 / 26 & 22 / 19 & 13 / 14 \\
 & Depictive & 103 & 68 / 72 & 41 / 18 & 43 / 18 & 12 / 10 & 7 / 7 \\
 & Purpose & 13 & 67 / 52 & 50 / 42 & 50 / 40 & 40 / 42 & 35 / 42 \\
 & Negation & 38 & 11 / 22 & 0 / 0 & 0 / 0 & 0 / 0 & 0 / 0 \\
\midrule
\multirow{4}{*}{LLaVA-OV2-8B} & Hedged & 190 & 88 / 72 & 51 / 39 & 52 / 42 & 25 / 26 & 8 / 12 \\
 & Depictive & 130 & 77 / 72 & 34 / 25 & 35 / 25 & 12 / 16 & 6 / 9 \\
 & Purpose & 62 & 76 / 40 & 21 / 16 & 22 / 16 & 13 / 11 & 8 / 7 \\
 & Negation & 39 & 58 / 36 & 8 / 3 & 9 / 5 & 3 / 0 & 1 / 0 \\
\bottomrule
\end{tabular}
\caption{\Fm{} FPR (\%) per category as \fm{} mentions are added to the training data. Cells: Lookback Lens /
mid-layer hidden-state probe. Columns as in Table~\ref{tab:mixing-side} (\emph{max} is 100\% for Qwen2.5-VL-7B and
LLaVA-OneVision-2). $n$: held-out mentions of the category per seed; cells with $n<30$ are noisy.}
\label{tab:cat4-mixing-sub}
\end{table*}




\section{Treating Hedges as Claims}
\label{app:hedgeclaim}
Table~\ref{tab:hedgeclaim} repeats the comparison of \S\ref{sec:res-findings} on all five models with every hedge
re-labelled as a claim.

\begin{table*}[t]
\centering\small\setlength{\tabcolsep}{2.5pt}
\resizebox{\textwidth}{!}{%
\begin{tabular}{l|cccc|cccc|cccc|cccc|cccc}
\toprule
 & \multicolumn{4}{c|}{LLaVA-1.5-7B} & \multicolumn{4}{c|}{Qwen2.5-VL-7B} & \multicolumn{4}{c|}{Qwen2.5-VL-3B} & \multicolumn{4}{c|}{InternVL3-8B} & \multicolumn{4}{c}{LLaVA-OV2-8B} \\
 & \multicolumn{2}{c}{FPR $\downarrow$} & \multicolumn{2}{c|}{TPR $\uparrow$} & \multicolumn{2}{c}{FPR $\downarrow$} & \multicolumn{2}{c|}{TPR $\uparrow$} & \multicolumn{2}{c}{FPR $\downarrow$} & \multicolumn{2}{c|}{TPR $\uparrow$} & \multicolumn{2}{c}{FPR $\downarrow$} & \multicolumn{2}{c|}{TPR $\uparrow$} & \multicolumn{2}{c}{FPR $\downarrow$} & \multicolumn{2}{c}{TPR $\uparrow$} \\
Detector & Ours & Claim & Ours & Claim & Ours & Claim & Ours & Claim & Ours & Claim & Ours & Claim & Ours & Claim & Ours & Claim & Ours & Claim & Ours & Claim \\
\midrule
\multicolumn{21}{l}{\textit{No non-assertive training data (0\%)}} \\
Lookback Lens       & 27 & 37 & 54 & 50 & 63 & 50 & 64 & 58 & 53 & 42 & 64 & 57 & 54 & 48 & 55 & 54 & 74 & 58 & 71 & 67 \\
Hidden state (mid)  & 46 & 58 & 49 & 49 & 57 & 51 & 69 & 62 & 40 & 29 & 64 & 55 & 58 & 56 & 63 & 63 & 57 & 51 & 75 & 73 \\
Hidden state (last) & 25 & 41 & 46 & 44 & 60 & 50 & 72 & 64 & 44 & 32 & 64 & 55 & 46 & 43 & 62 & 60 & 61 & 50 & 77 & 72 \\
DHCP                & 12 & 30 & 43 & 46 & 51 & 47 & 59 & 53 & 53 & 42 & 56 & 50 & 53 & 43 & 53 & 47 & 61 & 46 & 62 & 51 \\
Text-only LM        & 41 & 48 & 35 & 35 & 55 & 45 & 54 & 49 & 39 & 31 & 54 & 43 & 50 & 50 & 50 & 50 & 57 & 49 & 68 & 61 \\
\midrule
\multicolumn{21}{l}{\textit{Non-assertive training data at the natural share}} \\
Lookback Lens       & 9 & 15 & 54 & 50 & 24 & 18 & 58 & 58 & 17 & 12 & 64 & 55 & 32 & 22 & 54 & 56 & 26 & 22 & 68 & 67 \\
Hidden state (mid)  & 22 & 25 & 50 & 49 & 18 & 17 & 65 & 59 & 13 & 13 & 61 & 53 & 15 & 17 & 60 & 62 & 19 & 21 & 71 & 72 \\
Hidden state (last) & 12 & 20 & 43 & 43 & 17 & 18 & 66 & 61 & 13 & 13 & 60 & 57 & 16 & 14 & 59 & 59 & 20 & 21 & 70 & 70 \\
DHCP                & 6 & 17 & 37 & 40 & 28 & 20 & 60 & 49 & 20 & 13 & 56 & 48 & 20 & 19 & 47 & 46 & 37 & 33 & 58 & 58 \\
Text-only LM        & 15 & 21 & 35 & 35 & 13 & 15 & 52 & 44 & 15 & 15 & 50 & 43 & 13 & 12 & 48 & 51 & 19 & 20 & 62 & 61 \\
\bottomrule
\end{tabular}}
\caption{The failure does not depend on how hedges are labelled. \emph{Ours}: hedges are \fm{} (our convention);
\emph{Claim}: every hedge is re-labelled as a claim (object absent $\rightarrow$ hallucination, present
$\rightarrow$ concrete) and the probes are retrained. FPR (\%): held-out depictive, purpose and negation mentions
flagged as hallucinated (the same mentions under both conventions); TPR (\%): held-out hallucinations flagged.
Threshold flags 10\% of concrete mentions; each model on its own captions; mean of 3 seeds.}
\label{tab:hedgeclaim}
\end{table*}

\section{Concept Distribution}
\label{app:concepts}
Table~\ref{tab:concepts} lists the COCO concepts behind the \fm{} mentions of each model and category. Negation is
concentrated on a few concepts, mostly \emph{person} and \emph{car} (``no people or vehicles are visible''), whereas
hedged and depictive mentions span 41--71 concepts.

\section{Excluded Non-Assertive Mentions}
\label{app:other}

We exclude two types of candidates from \bench{}.
Compound expressions in which the matched word does not
refer to the object (e.g., ``hot dog'', ``teddy bear'')
account for 25--37\% of mentions initially labelled
\fm{}. Another 6--13\% are valid \fm{} mentions outside
our four subtypes, mostly generic statements (e.g.,
``a road where cars usually drive'').

We evaluate both excluded groups using the same detectors,
splits and thresholds as in the main experiments
(Table~\ref{tab:other}). Valid excluded mentions (Other)
exhibit similar detection failures: on the four newer
models, trained probes flag 36--75\% of them, compared
with 48--80\% for the four retained subtypes. SVAR flags
61--70\% of Other mentions, versus 46--57\% for the
retained subtypes. Compound matches are also frequently
flagged by trained probes, despite not referring to the
target object. These results show that the false-positive
problem extends beyond the four subtypes in \bench{}.

\begin{table*}[t]
\centering
\small
\setlength{\tabcolsep}{5pt}
\begin{tabular}{lccccccc}
\toprule
& \multicolumn{3}{c}{Trained Probes (\%) $\downarrow$}
& \multicolumn{3}{c}{SVAR (\%) $\downarrow$}
& Other $n$ \\
\cmidrule(lr){2-4}
\cmidrule(lr){5-7}
Model & 4 cat. & Other & Comp. & 4 cat. & Other & Comp. & per seed \\
\midrule
LLaVA-1.5-7B         & 18--52 & 14--32 & 11--39 & 57 & 67 & 19 & 282 \\
Qwen2.5-VL-7B        & 56--67 & 36--50 & 51--74 & 53 & 70 & 35 & 158 \\
Qwen2.5-VL-3B        & 48--66 & 40--56 & 47--62 & 53 & 65 & 30 & 166 \\
InternVL3-8B         & 53--62 & 56--75 & 52--60 & 46 & 61 & 22 & 46 \\
LLaVA-OV2-8B         & 60--80 & 53--69 & 61--77 & 57 & 63 & 42 & 95 \\
\bottomrule
\end{tabular}
\caption{Detector flag rates (\%, $\downarrow$) without
\fm{} training. \emph{4 cat.}: retained \fm{} subtypes;
\emph{Other}: excluded valid \fm{} mentions;
\emph{Comp.}: invalid compound matches.
Trained probe results show ranges across five detectors.
Same splits and thresholds as Table~\ref{tab:main};
mean over three seeds. $n$ denotes held-out Other mentions per seed.}
\label{tab:other}
\end{table*}
\begin{table*}[t]
\centering\small\setlength{\tabcolsep}{4pt}
\begin{tabular}{ll rr l}
\toprule
Model & Category & $n$ & Concepts & Five most frequent concepts (\% of the category) \\
\midrule
\multirow{4}{*}{LLaVA-1.5-7B} & Hedged & 564 & 58 & person 31, bird 7, bed 6, sports ball 5, couch 5 \\
 & Depictive & 265 & 50 & person 28, dog 11, train 6, bird 6, cat 6 \\
 & Purpose & 630 & 26 & person 48, sports ball 26, train 8, bus 4, tv 2 \\
 & Negation & 135 & 6 & person 69, car 27, boat 2, tv 1, train 1 \\
\midrule
\multirow{4}{*}{Qwen2.5-VL-7B} & Hedged & 958 & 69 & person 10, bed 7, cell phone 7, train 6, couch 5 \\
 & Depictive & 642 & 57 & person 23, bird 16, car 8, bicycle 7, dog 6 \\
 & Purpose & 182 & 36 & person 21, train 8, sports ball 7, cell phone 7, car 6 \\
 & Negation & 84 & 7 & person 61, car 21, kite 12, sports ball 2, tie 1 \\
\midrule
\multirow{4}{*}{Qwen2.5-VL-3B} & Hedged & 802 & 70 & person 12, bed 11, couch 7, oven 5, train 4 \\
 & Depictive & 516 & 49 & person 29, bird 11, car 7, dog 6, bear 5 \\
 & Purpose & 211 & 42 & person 28, cell phone 8, skis 7, sports ball 5, kite 5 \\
 & Negation & 216 & 8 & person 81, car 13, kite 3, train 1, boat $<$1 \\
\midrule
\multirow{4}{*}{InternVL3-8B} & Hedged & 198 & 45 & oven 12, person 8, bus 7, train 7, bed 6 \\
 & Depictive & 307 & 41 & person 27, bird 12, dog 7, car 6, bicycle 6 \\
 & Purpose & 43 & 15 & sports ball 21, skis 14, bed 12, person 9, keyboard 7 \\
 & Negation & 130 & 2 & person 89, car 11 \\
\midrule
\multirow{4}{*}{LLaVA-OV2-8B} & Hedged & 561 & 71 & person 10, bed 9, couch 8, train 7, bus 6 \\
 & Depictive & 450 & 51 & person 28, bird 13, car 8, dog 6, bicycle 4 \\
 & Purpose & 172 & 28 & person 27, sports ball 10, train 8, cell phone 7, skis 6 \\
 & Negation & 131 & 6 & person 81, car 15, kite 2, boat 2, bicycle 1 \\
\bottomrule
\end{tabular}
\caption{COCO concepts of the \fm{} mentions in \bench{}, per model and category: number of mentions ($n$, as in
Table~\ref{tab:dataset_counts}), number of distinct concepts, and the five most frequent concepts with their share of
the category (\%). The concept ``orange'' is excluded.}
\label{tab:concepts}
\end{table*}

\end{document}